\documentclass[sigconf]{acmart}

\AtBeginDocument{%
  }
    
\usepackage{multirow}
\usepackage{times}
\usepackage{array}
\usepackage{amsmath}
\usepackage{threeparttable}
\usepackage{makecell}
\usepackage{subfigure}
\usepackage{tabularx}
\usepackage{color}
\setcopyright{acmlicensed}
\copyrightyear{2025}
\acmYear{2025}
\acmDOI{XXXXXXX.XXXXXXX}

\acmConference[Conference acronym 'XX]{Make sure to enter the correct
  conference title from your rights confirmation emai}{June 03--05,
  2025}{Woodstock, NY}
\acmISBN{978-1-4503-XXXX-X/18/06}

\begin{document}

\title{MM-Path: Multi-modal, Multi-granularity Path Representation Learning---Extended Version}


\settopmatter{authorsperrow=4} 
\author{Ronghui Xu}
\affiliation{%
  \institution{East China Normal University}
  \city{Shanghai}
  \country{China}
}
\email{rhxu@stu.ecnu.edu.cn}

\author{Hanyin Cheng}
\affiliation{%
  \institution{East China Normal University}
  \city{Shanghai}
  \country{China}
}
\email{hycheng@stu.ecnu.edu.cn}

\author{Chenjuan Guo}
\affiliation{%
  \institution{East China Normal University}
  \city{Shanghai}
  \country{China}
}
\email{cjguo@dase.ecnu.edu.cn}

\author{Hongfan Gao}
\affiliation{%
  \institution{East China Normal University}
  \city{Shanghai}
  \country{China}
}
\email{hf.gao@stu.ecnu.edu.cn}

\author{Jilin Hu}
\affiliation{%
  \institution{East China Normal University}
  \city{Shanghai}
  \country{China}
}
\affiliation{%
  \institution{KLATASDS-MOE}
  \city{Shanghai}
  \country{China}
}
\email{jlhu@dase.ecnu.edu.cn}

\author{Sean Bin Yang}
\affiliation{%
  \institution{Chongqing University of Posts and Telecommunications}
  \city{Chongqing}
  \country{China}
}
\email{sean.byang@gmail.com}

\author{Bin Yang} 
\authornote{Corresponding authors}
\affiliation{%
  \institution{East China Normal University }
  \city{Shanghai}
  \country{China}
}
\email{byang@dase.ecnu.edu.cn} 

\renewcommand{\shortauthors}{Ronghui Xu et al.}

\begin{abstract}
  Developing effective path representations has become increasingly essential across various fields within intelligent transportation. Although pre-trained path representation learning models have shown improved performance, they predominantly focus on the topological structures from single modality data, i.e., road networks, overlooking the geometric and contextual features associated with path-related images, e.g., remote sensing images. Similar to human understanding, integrating information from multiple modalities can provide a more comprehensive view, enhancing both representation accuracy and generalization. However, variations in information granularity impede the semantic alignment of road network-based paths (road paths) and image-based paths (image paths), while the heterogeneity of multi-modal data poses substantial challenges for effective fusion and utilization.  In this paper, we propose a novel Multi-modal, Multi-granularity Path Representation Learning Framework (MM-Path), which can learn a generic path representation by integrating modalities from both road paths and image paths. To enhance the alignment of multi-modal data, we develop a multi-granularity alignment strategy that systematically associates nodes, road sub-paths, and road paths with their corresponding image patches, ensuring the synchronization of both detailed local information and broader global contexts.  To address the heterogeneity of multi-modal data effectively, we introduce a graph-based cross-modal residual fusion component designed to comprehensively fuse information across different modalities and granularities. Finally, we conduct extensive experiments on two large-scale real-world datasets under two downstream tasks, validating the effectiveness of the proposed MM-Path. This is an extended version of the paper accepted by KDD 2025. The code is available at: \href{https://github.com/decisionintelligence/MM-Path}{https://github.com/decisionintelligence/MM-Path}.
\end{abstract}




\maketitle

\section{Introduction}
Understanding paths and developing effective path representations are increasingly essential, offering invaluable insights for diverse fields such as intelligent navigation~\cite{guo2018learning, yang2018pace, guo2024efficient, Simon2023, DBLP:journals/pvldb/PedersenYJ20}, route recommendation~\cite{dai2015personalized, zhang2022multi, chen2018price},  urban planning~\cite{xu2023, chen2021robust,DBLP:journals/sigmod/GuoJ014}, and urban emergency management~\cite{elbery2020iot}. Recent studies focus on developing pre-trained path representation learning models, which have demonstrated outstanding generalization capabilities~\cite{yang2023,  jiang2023start,chang2023contrastive}. These models efficiently produce generic path representations in an unsupervised manner. With simple fine-tuning and little labeled data, they are adaptable to diverse downstream tasks such as travel time estimation and path ranking score estimation. Consequently, they significantly improve computational efficiency by reducing both labeled data and runtime.

 \begin{figure}[!tp]
	\begin{center}
\includegraphics[width=0.88\linewidth]{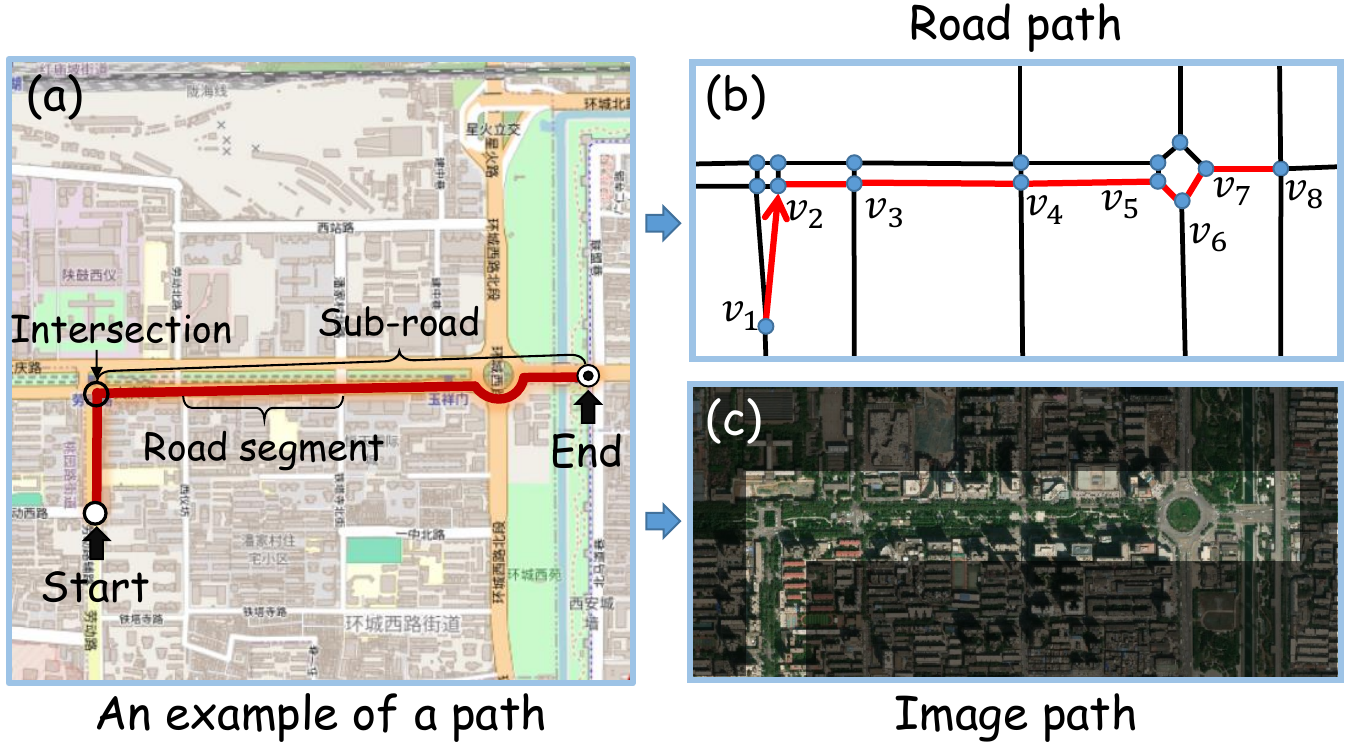}
	\end{center}
	\caption{A path in different modalities}
	\label{fig:intro}
\end{figure}

Paths have different modalities that provide richer, more diverse information. For example, while paths derived from road networks (road paths for short) elucidate topological relationships among road segments in paths, remote sensing images of paths (image paths for short) provide insights into geometric features and broader environmental contexts (see Figure~\ref{fig:intro}). Integrating these modalities enriches path representations with varied perspectives, thereby improving accuracy and enhancing generalization capabilities. However, current path representation learning models primarily rely on single-modality data from road networks, which fails to capture the deep, comprehensive context essential for a complete understanding of paths. This calls for developing a multi-modal pre-trained path representation learning model. Nonetheless, constructing such a model faces several challenges:

\textbf{Information granularity discrepancies between road paths and image paths significantly hinder cross-modal semantic alignment. }  
Effective cross-modal alignment, which ensures semantic consistency and complementarity among various modalities, is crucial for constructing multi-modal models~\cite{liang2024foundations}. However, the discrepancies in information granularity between road paths and image paths are substantial. As depicted in Figure~\ref{fig:intro}, road paths typically focus on detailed topological structures and delineate road connectivity, while the image paths capture global environmental contexts on a large scale, reflecting the functional attributes of corresponding regions. It is worth noting that images may include extensive regions that show low relevance to the road paths, such as the dark regions in Figure~\ref{fig:intro} (c). 
Current image-text multi-modal models~\cite{radford2021learning, bao2022vlmo, wang2023image,menon2022visual} typically align individual images with textual sequences. However, such single-granularity and coarse alignment methods introduce noise, which are not suitable for the precise alignment required for paths.
Additionally, as shown in Figure~\ref{fig:intro} (a), roads have different granularities in nature, including intersections, road segments, and sub-roads. Fully understanding paths at different granularities can provide insights from micro to macro levels, mitigating the negative effects caused by the differences in information granularity across modalities. Although some studies~\cite{chen2024pathformer, pan2023magicscaler} have explored multi-granularity in single-modal data, they have not adequately addressed the requirements for multi-granularity analysis in multi-modal contexts.
Thus, it is crucial to refine multi-granularity data processing and develop multi-granularity methods for cross-modal alignment.

\textbf{The inherent heterogeneity of road paths and image paths poses a significant challenge during feature fusion.} 
The differences in data structure and information granularity between road paths and image paths extend to their learning methods. Road path representation learning typically focuses on connectivity and reachability between roads and intersections, as well as analyzing graph structures~\cite{yang2022,yang2020context,yang2023,jiang2023start}. Conversely, image learning methods that are able to learn image paths prioritize object recognition and feature extraction, aiming for a broad understanding of image content~\cite{bao2021beit,he2016deep}.
These disparate learning methods lead to road paths and image paths mapped to different embedding spaces, resulting in feature dimensions with similar semantics containing entirely different information. Simple fusion methods like early fusion (i.e., integrating multiple modalities before or during the feature extraction stage) and late fusion (i.e., keeping each modality independently processed until the final fusion stage) may result in information loss and increased bias, and fail to capture subtle correlations between road paths and image paths~\cite{yang2023code, liang2024foundations}. Therefore, a multi-modal fusion method that can capture the relationships among entities in different modalities and ensuring effective data fusion, is critically needed.

To address these challenges, we propose a \textit{Multi-modal, Multi-granularity Path Representation Learning Framework}, namely MM-Path, for learning generic path representations. 

\textbf{To address the first challenge}, we propose a multi-granularity alignment component. This component systematically associates intersections, road sub-paths, and entire road paths with their corresponding image information to capture details accurately at a finer granularity as well as maintaining global correspondence at a coarser granularity.
Specifically, we divide the image of the entire interested region into small fixed-size images, collect the small fixed-size images along each path, and arrange the collected images into an image path (i.e., image sequence). 
We employ modal-specific tokenizers to generate the initial embeddings for road paths and image paths, respectively. Subsequently, these initial embeddings are fed into the powerful Transformer architecture to learn complex encoded embeddings for each modality at three granularities.
Finally, a multi-granularity alignment loss function is employed to ensure the alignment of road and image encoded embeddings across different granularities.

\textbf{To address the second challenge}, we introduce a graph-based cross-modal residual fusion component, which is designed to effectively fuse cross-modal features while incorporating spatial contextual information. Specifically, we link the encoded embeddings of each modality with the initial embeddings of the other modality to create road and image residual embeddings, respectively, with the purpose of fusing cross-modal features from different stages. We then build a cross-modal adjacency matrix for each path based on spatial correspondences and contextual information. This matrix guides the GCN to iteratively fuse the residual embeddings of each modality separately, thus obtaining road and image fused embeddings. Finally, we apply contrastive loss to ensure the consistency of the fused embeddings across the two modalities. As the final representation effectively integrates cross-stage features of the two modalities with spatial context information, this component not only achieves deep multi-modal fusion but also enhances the comprehensive utilization of information.

The contributions of this work are delineated as follows: 
\begin{itemize}
    \item We propose a Multi-modal, Multi-granularity Path Representation Learning Framework that learns generic path representations applicable to various downstream tasks. To the best of our knowledge, MM-Path is the first model that leverages road network data and remote sensing images to learn generic path representations.
    
    \item We model and align the multi-modal path information using a fine-to-coarse multi-granularity alignment strategy. This strategy effectively captures both intricate local details and the broader global context of the path.

    \item We introduce a graph-based cross-modal residual fusion component. This component utilizes a cross-modal GCN to fully integrate information from different modalities while maintaining the consistency of dual modalities.
    
    \item We conduct extensive experiments on diverse tasks using two real-world datasets to demonstrate the adaptability and superiority of our model.
\end{itemize}

\section{Preliminaries}

\subsection{Basic Conception}
\textbf{Path.} A path $p$ is a sequence of continuous junctions, which can be observed from the road network view and the image view.

\noindent
\textbf{Road network.} A road network is denoted as $\mathcal{G}=(\mathcal{V},\mathcal{E})$, where $\mathcal{V}$ and $\mathcal{E}$ represent a set of nodes and edges, respectively. Node $v \in \mathcal{V}$ is a road intersection or a road end. Edge $e \in \mathcal{E}$ denotes a road segment connecting two nodes.

\noindent
\textbf{Road paths.} We define the sequence of nodes on a road network for a path $p$ as a road path $\mathcal{R}(p)=\langle v_1,v_2,...,v_{|\mathcal{R}(p)|}\rangle$, where each element represents a node, and $|\mathcal{R}(p)|$ represents the length of the road path $\mathcal{R}(p)$. It is noted that there must be an edge $e \in \mathcal{E}$ connecting any adjacent nodes in the road path.

\noindent
\textbf{Image paths.} 
Given an interested region, we partition the region into fixed-size segments to generate a set of images, $\mathcal{M}_{rs}$, consisting of  disjoint, fixed-size remote sensing images.
Each image within this set is denoted as $m \in \mathbb{R}^{r \times r \times c}$, where $c$ represents the number of channels, and $(r,r)$ denotes the resolution. Subsequently, given road path $\mathcal{R}(p)$, the image path (i.e., image sequence of the path) $\mathcal{M}({p})$ is formed by selecting a series of image $m_i$ that correspond to specific latitudes and longitudes along the nodes in the road path. 

\begin{figure}[!ht]
	\begin{center}
\includegraphics[width=0.8\linewidth]{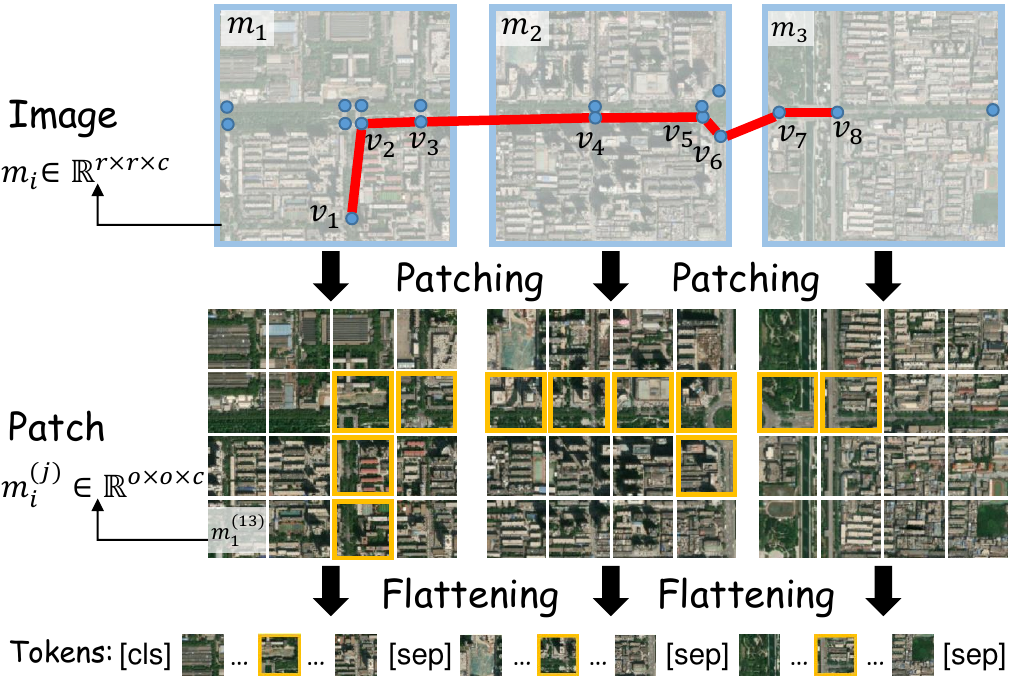}
	\end{center}
	\caption{An example of image path processing} 
\label{fig:input}
\end{figure}

For example, as shown in the upper part of Figure~\ref{fig:input}, consider the road path $\mathcal{R}(p)=\langle v_1,...,v_{8}\rangle$, where nodes $v_1, v_2$, and $v_3$ are located in image $m_1$, nodes $v_4, v_5$ and $v_6$ in image $m_2$, and nodes $v_7$ and $v_8$ in image $m_3$. This results in the image path $\mathcal{M}({p})=\langle m_1,m_2,m_{3}\rangle$.

\noindent
\textbf{Road Sub-paths.} Given a road path $\mathcal{R}({p})$ and an image path $\mathcal{M}({p})$, the nodes of $\mathcal{R}({p})$ located in the same image belong to a road sub-path. 

Taken Figure~\ref{fig:input} as an example, the road path $\mathcal{R}({p})=\langle v_1,\dots,v_{8}\rangle$ has three road sub-paths: $s_1=\langle v_1,v_2,v_3\rangle$, $s_2=\langle v_4,v_5,v_6 \rangle$ and $s_3=\langle v_7, v_8\rangle$.

\subsection{Problem Statement}
Given the road path $\mathcal{R}({p})$ and image path $\mathcal{M}({p})$ of a path $p$, the goal is to learn an embedding function $f$ that returns a generic representation of path $p$. This function can be formalized as follows: 
\begin{equation}
\begin{aligned}
\mathbf{x}=f(\mathcal{R}({p}),\mathcal{M}({p})),
\end{aligned}
\end{equation}
where $\mathbf{x} \in \mathbb{R}^{d}$ represents the generic embedding of path $p$, and $d$ denotes the dimension of the embedding $\mathbf{x}$. 

These learned path embeddings are supposed to be generic, which should support a variety of downstream tasks, e.g., path travel time estimation and path ranking score estimation.

\begin{figure*}[tp]
	\centering
 \includegraphics[width=0.98\linewidth]{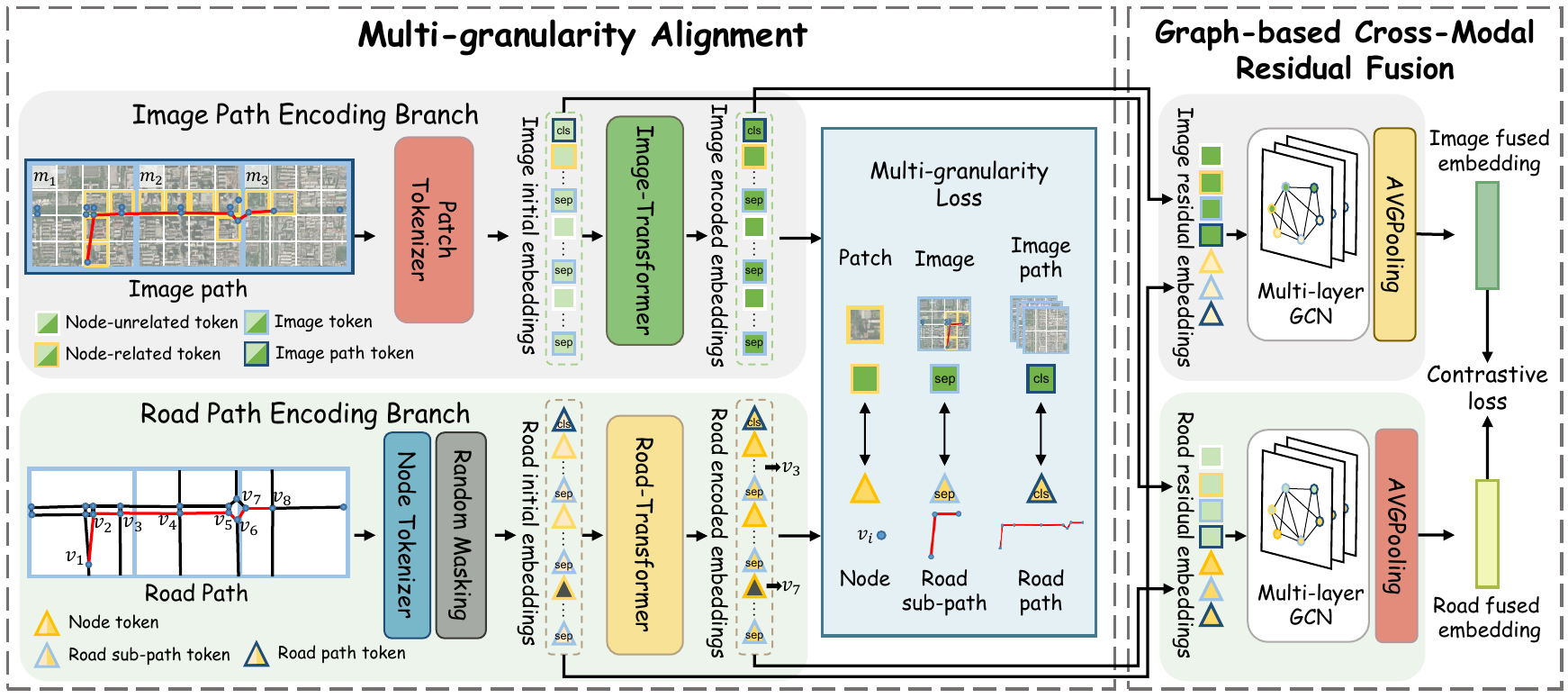}
  \caption{Overall framework of MM-Path }
 \label{fig:framework}
\end{figure*}

\section{Methodology}

Figure~\ref{fig:framework} illustrates the framework of MM-Path. This section introduces  the framework of MM-Path, describes its two main components, and details the final training objective.

\subsection{Overall Framework}
Different from existing methods that are limited to data from a single modality, MM-Path leverages data from both road networks and images for pre-training, providing a more comprehensive perspective. MM-Path comprises two main components: the multi-granularity alignment component and the graph-based cross-modal  residual fusion component.   

The multi-granularity alignment component is designed to concentrate on path-related information while capturing fine-grained details and coarse-grained global context. Initially, we convert the image of interested region into fixed-sized image sequences to obtain image paths. Subsequently, We establish road path and image path encoding branches to process the two modalities. In each branch, a modal-specific tokenizer generates initial embeddings for each modality at three granularities: node/patch, road sub-path/image, and road path/image path. These initial embeddings are then processed by road and image transformers to produce road and image encoded embeddings, respectively, which are also generated at the same three granularities. A multi-granularity loss function is utilized to synchronize the semantic information of road encoded embeddings and image encoded embeddings, and to capture their interrelations at different granularities, from fine to coarse.

The graph-based cross-modal residual fusion component is designed to effectively fuse cross-modal heterogeneous data. A cross-modal residual connection merges the encoded embedding from each branch with the initial embedding from the other branch, generating road and image residual embeddings. This connection considers cross-modal features at different stages, promoting deep cross-modal feature fusion.
Subsequently, we construct a cross-modal adjacency matrix for each path based on spatial correspondences and contextual information. This matrix, embedded within a GCN, guides the fusion of the two modalities for each branch. Consequently, a fused embedding is obtained for each branch. We introduce a contrastive loss to ensure the consistency between the road fused embedding and image fused embedding. Finally, we concatenate these two fused embeddings to obtain a generic path representation.

\subsection{Multi-granularity Alignment}

We model the road paths and image paths using Transformer architecture, respectively. We then construct a multi-granularity loss function to ensure alignment between these two modalities.

\subsubsection{Input Representations}
Due to information granularity discrepancies between road paths and image paths, direct alignment is often disturbed by irrelevant information. To solve this problem, we use a sequence of fixed-size images, instead of a single image commonly used in traditional image-text multi-modal methods~\cite{radford2021learning, bao2022vlmo, wang2023image,menon2022visual}, to model image paths.
This procedure preserves the scale and shape features of the images by avoiding distortions caused by inconsistencies in image sizes. Furthermore, as these fixed-size images can be utilized across different paths, the storage of images is reduced. Then, we utilize specialized tokenizers to separately encode the data of each modality into a unified format.

The patch tokenizer segments each image within an image path into a series of patches to extract fine-grained semantic information. Specifically, as shown in Figure~\ref{fig:input}, an image $m_i \in \mathbb{R}^{r \times r \times c}$ is reshaped into a sequence of ${r^2/o^2}$ (e.g., 16) patches, where $c$ represents the number of channels, $(r,r)$ denotes the resolution of fixed-size image, and $(o, o)$ defines the resolution per patch. After patching, we concatenate the patch sequences from all images within a image path to form a unified patch sequence. Then, we place a special [cls] token at the beginning of the patch sequence. As the [cls] token captures the global information of the entire sequence~\cite{bao2021beit}, it can be regarded as a representation of the entire image path. Special [sep] tokens are placed at the end of each image to delineate local information of each image. For example, the token sequence of image path $\mathcal{M}({p})=\langle m_1,m_2,m_3\rangle$ is $[\text{\text{cls}}, m_{1}^{(1)}, \dots, m_{1}^{(16)}, \text{sep},\dots,\text{sep}, \mathbf{m}_{3}^{(1)}, \dots, m_{3}^{(16)},\text{sep}]$, where $m_{i}^{(j)} \in \mathbb{R}^{o\times o \times c}$ denotes the $j$-th patch of the $i$-th image.

Each patch $m_{i}^{(j)}$ is then projected into a patch embedding $\mathbf{m}_{i}^{(j)} \in \mathbb{R}^{d}$, which can be initialized using pre-trained ResNet50~\cite{he2016deep}.
The image initial embeddings are computed by summing the patch embeddings with the image position embeddings $\mathbf{T}_\text{image} \in \mathbb{R}^{n_1 \times d}$, resulting in $\mathbf{H}^{(0)} = [\mathbf{m}_{\text{cls}}, \mathbf{m}_{1}^{(1)}, ..., \mathbf{m}_{|\mathcal{M}({p})|}^{(r^2/o^2)},  \mathbf{m}_{\text{sep}}] + \mathbf{T}_\text{image}$. Here, $\mathbf{m}_{\text{cls}}$ and $\mathbf{m}_{\text{sep}}$ are the image initial embeddings of the [cls] and [sep] tokens, respectively. $n_1$ denotes the length of the patch token sequence, and $d$ represents the dimension of the embeddings.

The modeling for a road path is similar, starting with a [cls] token at the beginning and placing [sep] tokens at the end of each road sub-path.  For instance, a road path $\mathcal{R}({p})$ comprising three road sub-paths---$s_1=\langle v_1,v_2,v_3\rangle$, $s_2=\langle v_4,v_5,v_6\rangle$, and $s_3=\langle v_7,v_8\rangle$---generates the node token sequence $[\text{cls}, v_1, v_2, v_3, \text{sep}, v_4, v_5,v_6, \text{sep}, v_7, \\ v_8,\text{sep}]$.
Then, the node tokenizer linearly projects each node $v_i$ from the road path $\mathcal{R}({p})$ to a node embedding $\mathbf{v}_i \in \mathbb{R}^{d}$, initialized using Node2vec~\cite{grover2016node2vec}. 
We also integrate standard learnable road position embeddings $\mathbf{T}_\text{road} \in \mathbb{R}^{n_2 \times {d}}$ with the node embeddings to generate the road initial embeddings $\mathbf{P}^{(0)}=[\mathbf{v}_{\text{cls}}, \mathbf{v}_{1}, ..., \mathbf{v}_{|\mathcal{R}({p})|}, \mathbf{v}_{\text{sep}}]+\mathbf{T}_\text{road}$, where $\mathbf{v}_{\text{cls}}$ and $\mathbf{v}_{\text{sep}}$ represent the road initial embeddings of the [cls] and [sep] tokens, respectively. $n_2$ denotes the length of the road path token sequence.

\subsubsection{Image Path Encoding}
To model the image path, the image initial embeddings $\mathbf{H}^{(0)}$ are fed into the $l$ layers Image-Transformer, which can be formulated as,
\begin{equation}
\begin{aligned}
\mathbf{H}^{(j)}=\text{Image-Transformer}(\mathbf{H}^{(j-1)}),
\end{aligned}
\end{equation}
where $j=1,...,l$, and $\mathbf{H}^{(l)}  \in \mathbb{R}^{n_1 \times d}$ is the final output of the Image-Transformer. For a simplify, we denote the image encoded embeddings $\mathbf{H}^{(l)}$ as $\mathbf{H}=[\mathbf{h}_{\text{cls}}, \mathbf{h}_{1}^{(1)}, ..., \mathbf{h}_{|\mathcal{M}({p})|}^{(r^2/o^2)},  \mathbf{h}_{\text{sep}_{|\mathcal{M}({p})|}}]$. $\mathbf{h}_{i}^{(j)} \in \mathbb{R}^{d}$ denotes the encoded embedding of the $j$-th patch of the $i$-th image. $\mathbf{h}_{\text{sep}_{i}}, \mathbf{h}_{\text{cls}} \in \mathbb{R}^{d}$ represent the encoded embeddings of the $i$-th image and the whole image path $\mathcal{M}({p})$, respectively.

\subsubsection{Road Path Encoding}

To facilitate alignment between road paths and image paths, we utilize a similar Transformer architecture for road path modeling. The encoder is comprised of $l$ layers of Transformer blocks, defined as:
\begin{equation}
\begin{aligned}
\mathbf{P}^{(j)}=\text{Road-Transformer}(\mathbf{P}^{(j-1)}),
\end{aligned}
\end{equation}
where $j=1,...,l$, $\mathbf{P}^{(j)} \in \mathbb{R}^{n_2 \times d}$ is the output of the $j$-th layer. In a brief, the road encoded embeddings $\mathbf{P}^{(l)}$ are represented as $\mathbf{P}=[\mathbf{p}_{\text{cls}}, \mathbf{p}_{1},\dots, \mathbf{p}_{|\mathcal{R}(p)|},\mathbf{p}_{\text{sep}_{|\mathcal{M}({p})|}}]$. Here, $\mathbf{p}_{i}, \mathbf{p}_{\text{sep}_{i}}, \mathbf{p}_{\text{cls}} \in \mathbb{R}^{d}$ denote the encoded embeddings of the $i$-th node, the sub-path $s_i$, and the entire road path $\mathcal{R}(p)$, respectively. $|\mathcal{R}(p)|$ represents the number of nodes in road path $\mathcal{R}(p)$, and $|\mathcal{M}({p})|$ indicates the number of images in image path $\mathcal{M}({p})$, which also corresponds to the number of road sub-paths.

To better capture the complex dependencies in a path, similar to masked language modeling task (MLM)~\cite{kenton2019bert}, we use a masked node modeling task as a self-supervised task. The intuition behind this is that the information density of individual pixels or patches in an image is relatively low compared to the topological structure information relevant to the path. Therefore, image masking tasks are not deemed essential. To this end, we propose to employ node masking tasks. In particular,
we randomly mask the nodes in road paths (Ref. as to the gray triangle in Figure~\ref{fig:framework}) , and then use a softmax classifier to predict the node tokens corresponding to the masked nodes. The loss function for training is defined as follows: 
\begin{equation}
\begin{aligned}
\mathcal{L}_\text{mask}=-\sum_{p\in \mathcal{P}} \sum_{i\in \mathcal{D}} \log \text{P}(v_i|v_i^{\text{mask}}),
\end{aligned}
\end{equation}
where $\mathcal{P}$ represents the training sets of all paths, $\mathcal{D}$ is randomly masked positions of road path, and $v_i^{\text{mask}}$ is the node that is masked according to $\mathcal{D}$.

\subsubsection{Modalities Aligning}
The encoded embeddings from each branch capture the hidden semantic information within their respective modality, including fine-grained node/patch embeddings, medium-grained road sub-path/image embeddings, and coarse-grained entire road path/image path embeddings. We aim for embeddings with similar semantics across modalities to be proximate within the embedding space. Additionally, we seek detailed alignment between the two modalities while maintaining global correspondence. 
Accordingly, we design a loss function that operates at three distinct levels of granularity---fine, medium and coarse---corresponding to node/patch, road sub-path/image, and entire road path/image path, respectively.

\textbf{Fine granularity.}
Since each patch may contain more than one node, the encoded embeddings of a node and the corresponding patch (Ref. as to the dark yellow triangle the dark green rectangle with yellow borders in Figure~\ref{fig:framework}) should maintain directional consistency. To precisely capture the semantic information of fine-grained paths, we minimize the cosine distance between the encoded embeddings of nodes and their corresponding patches.  Consequently, we construct the fine-grained loss function as follows:
\begin{equation}
\begin{aligned}
\mathcal{L}_{\text{fine}}=\sum_{p \in \mathcal{P}} \sum_{v_i \in \mathcal{R}(p), \text{L}(v_i)=m_{j}^{(k)}} (1-\frac{\mathbf{p}_i \cdot \mathbf{h}_{j}^{(k)}}{\Vert \mathbf{p}_i \Vert \Vert \mathbf{h}_{j}^{(k)} \Vert}),
\end{aligned}
\end{equation}
where $\mathcal{P}$ represents the training set of paths, $\text{L}(v_i)$ is a function that returns the patch corresponding to the node $v_i$, $\mathbf{p}_i$ and $\mathbf{h}_{j}^{(k)}$ are the encoded embeddings of $v_i$ and $m_{j}^{(k)}$, respectively. 

\textbf{Medium granularity.} Similarly, to align road sub-paths with images, we construct the following medium-grained loss function:
\begin{equation}
\begin{aligned}
\mathcal{L}_{\text{medium}}=\sum_{p \in \mathcal{P}} \sum_{s_i \in \mathcal{R}(p)} (1-\frac{\mathbf{p}_{\text{sep}_i} \cdot\mathbf{h}_{\text{sep}_i}}{\Vert \mathbf{p}_{\text{sep}_i} \Vert \Vert \mathbf{h}_{\text{sep}_i} \Vert}),
\end{aligned}
\end{equation}
where $\mathbf{p}_{\text{sep}_i}$ and $\mathbf{h}_{\text{sep}_i}$ (Ref. as to the dark yellow rectangle and the dark green triangle with blue borders in Figure~\ref{fig:framework}) are the encoded embeddings of road sub-path $s_i$ and the corresponding image $m_i$, respectively.

\textbf{Coarse granularity.} 
Due to the unique correspondence between road path and the corresponding image path, a clearer distinction is necessary. Therefore, We construct a contrastive loss function for coarse-grained data.
Considering a batch of road path-image path pairs $\mathcal{B}$, the objective of this contrastive learning loss is to accurately identify the matched pairs among the $|\mathcal{B}| \times |\mathcal{B}|$ possible combinations. Within a training batch, there are $|\mathcal{B}|^2-|\mathcal{B}|$ negative pairs. The contrastive learning loss function can be formulated as:
\begin{equation}
\begin{aligned}
\mathcal{L}_{\text{coarse}}&=-\sum_{p \in \mathcal{P}} (\log(\frac{\exp (\text{sim}(\mathbf{p}_{\text{cls}},\mathbf{h}_{\text{cls}})/\sigma)}{\sum_{m^\text{Neg} \in \mathcal{B}} \exp (\text{sim}(\mathbf{p}_{\text{cls}},\mathbf{h}^\text{Neg}_{\text{cls}})/\sigma) }\\
&+\frac{\exp (\text{sim}(\mathbf{p}_{\text{cls}},\mathbf{h}_{\text{cls}})/\sigma)}{\sum_{p^\text{Neg} \in \mathcal{B}} \exp (\text{sim}(\mathbf{p}^\text{Neg}_{\text{cls}},\mathbf{h}_{\text{cls}})/\sigma) }),\\
\end{aligned}
\end{equation}
where $\mathbf{p}_{\text{cls}}$ and $\mathbf{h}_{\text{cls}}$ (Ref. as to the dark yellow rectangle and the dark green triangle with dark blue borders in Figure~\ref{fig:framework}) correspond to the encoded embeddings of the entire road path and image path in a road path-image path pairs. $p^\text{Neg}$ and $m^\text{Neg}$ are the negative road path and image path in the batch set $\mathcal{B}$, respectively. $\sigma$ is a learned temperature parameter. $\text{sim}(\mathbf{p}_{\text{cls}},\mathbf{h}_{\text{cls}})$ returns the Euclidean distance between $\mathbf{p}_{\text{cls}}$ and $\mathbf{h}_{\text{cls}}$. 

Finally, the multi-granularity loss can be formulated as $\mathcal{L}_{\text{multi}}= \mathcal{L}_\text{fine }+\mathcal{L}_\text{medium}+\mathcal{L}_\text{coarse}$.

\subsection{Graph-based Cross-modal Residual Fusion}

In this framework, each path is represented through two distinct modalities, providing complementary perspectives. To effectively leverage these modalities, we propose a graph-based cross-modal residual fusion component.

\subsubsection{Cross-modal Residual Connection}
To facilitate comprehensive information exchange between modalities, we introduce cross-modal residual connections that effectively concatenate embeddings across different stages and modalities. These connections enable direct propagation of gradients to earlier layers, thereby enhancing stability and improving training efficiency.
Specifically, we concatenate the road initial embeddings $\mathbf{P}^{(0)}$ with the image encoded embeddings $\mathbf{H}$, and the image initial embeddings $\mathbf{H}^{(0)}$ with the road encoded embeddings $\mathbf{P}$. The resulting image residual embeddings and road residual embeddings are defined as $\mathbf{U}=\mathbf{P}^{(0)} \Vert \mathbf{H}$ and $\mathbf{Q}=\mathbf{P}\Vert \mathbf{H}^{(0)}$, respectively. Here, $\mathbf{U}, \mathbf{Q} \in \mathbb{R}^{(n_1+n_2) \times d}$, and $\Vert$ denotes the concatenation operation.

\subsubsection{Graph-based Fusion}
Although traditional attention mechanisms proficiently identify correlations among entities, they often fail to incorporate contextual information concurrently. To address this limitation, we utilize graph neural networks~\cite{cao2015grarep}, which incorporate contextual information into the learning process by representing it as graph structures. Leveraging this capability, we introduce a graph-based fusion method to enhance the accuracy of information understanding across different modalities. 

Initially, we construct a specialized cross-modal directed graph for each path. This graph treats all tokens, including [cls] and [sep] tokens from both modalities, as entities. These entities are connected via three types of relationships: intra-modal context, cross-modal correspondence, and cross-modal context. The intra-modal context focuses on interactions within a single modality, facilitating a deep understanding of its specific information. Cross-modal correspondence aids in comprehending and learning the spatial correspondence between different modalities. Cross-modal context addresses indirect relationships between different modal entities, which enhances the model's ability to interpret complex scenes. Collectively, leveraging these relationships significantly boosts the model’s capacity to handle multi-modal data effectively.

\begin{figure}[!tp]
	\begin{center}
	\includegraphics[width=0.95\linewidth]{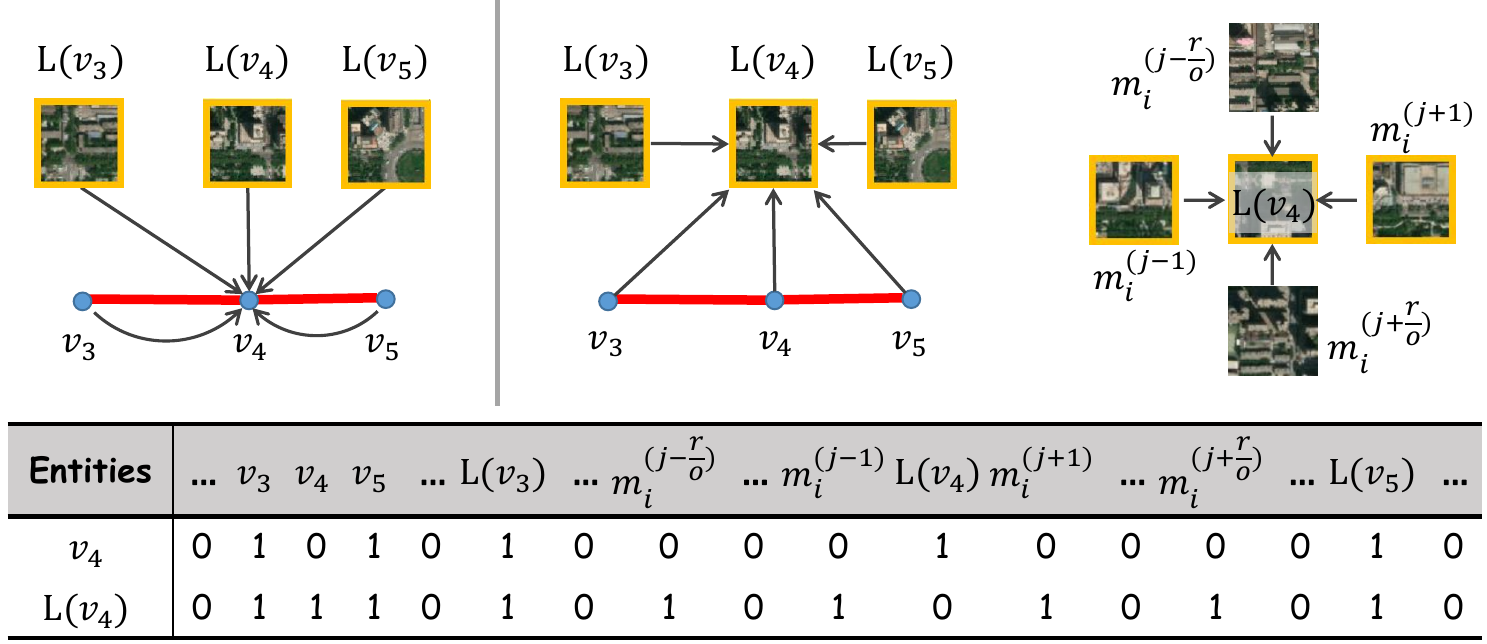}
	\end{center}
	\caption{An example of multi-modal graph construction}
  \label{fig:adjacent}
\end{figure}

Figure~\ref{fig:adjacent} demonstrates the construction of the graph. Taking node $v_4$ as an example, node $v_4$ is connected by directed edges from five entities: adjacent context nodes $v_3$ and $v_5$, its corresponding patch $\text{L}(v_4)$,  and their respective patches $\text{L}(v_3)$ and $\text{L}(v_5)$. The patch $\text{L}(v_4)$ (i.e., $m_{i}^{(j)}$) is connected by directed edges from nine entities, including $v_3$, $v_4$, $v_5$, $\text{L}(v_3)$, $\text{L}(v_5)$, and four geographically adjacent image patches $m_{i}^{(j-1)},m_{i}^{(j+1)},m_{i}^{(j-\frac{r}{o})}$ and $m_{i}^{(j+\frac{r}{o})}$, where $\frac{r}{o}$ denotes the number of patches per row of an image.
Additionally, the [sep] tokens are connected by their context [sep] tokens, corresponding [sep] tokens from another modality, and cross-modal context tokens. The [cls] token, encapsulating more global information, is connected by all [sep] tokens and corresponding [cls] token from another modality.

Then, we construct an adjacency matrix $\textbf{A} \in \mathbb{R}^{(n_1+n_2) \times (n_1+n_2)}$ for each path $p$ to capture the comprehensive relation within the multi-modal data. Given the effectiveness of GCNs in transferring and fusing information across entities within a graph structure, we employ a GCN to derive the updated embeddings for both branches. The updated embeddings are computed as follows:
\begin{equation}
\begin{aligned}
\hat{\mathbf{U}}=\text{Relu}\left(\tilde{\textbf{D}}^{-\frac{1}{2}} \tilde{\textbf{A}} \tilde{\textbf{D}}^{-\frac{1}{2}}\text{Relu}\left(\tilde{\textbf{D}}^{-\frac{1}{2}} \tilde{\textbf{A}} \tilde{\textbf{D}}^{-\frac{1}{2}} \mathbf{U} \textbf{W}_1 \right) \textbf{W}_2 \right),
\end{aligned}
\end{equation}
\begin{equation}
\begin{aligned}
\hat{\mathbf{Q}}=\text{Relu}\left(\tilde{\textbf{D}}^{-\frac{1}{2}} \tilde{\textbf{A}} \tilde{\textbf{D}}^{-\frac{1}{2}}\text{Relu}\left(\tilde{\textbf{D}}^{-\frac{1}{2}} \tilde{\textbf{A}} \tilde{\textbf{D}}^{-\frac{1}{2}} \mathbf{Q} \textbf{W}_3 \right) \textbf{W}_4 \right),
\end{aligned}
\end{equation}
where $\textbf{W}_1, \textbf{W}_2, \textbf{W}_3, \textbf{W}_4 \in \mathbb{R}^{d \times d}$ are weight matrices, and $\tilde{D}$ is the degree matrix of $\tilde{\textbf{A}}$.
The augmented adjacency matrix $\tilde{\textbf{A}}=\textbf{A}+\textbf{I}'$, where $\textbf{I}'$ is a modified identity matrix with all diagonal elements set to 1,  except for those corresponding to patches without relationship to any nodes (Ref. as to the dark green rectangle with a white border in Figure~\ref{fig:framework}). This modification aims to exclude patches that are relatively unrelated to the path, thereby preventing the introduction of noise into the model.

After iterative graph convolution operations, the embeddings of each entity within the graph are updated.  We perform average pooling on $\hat{\mathbf{U}}$ and $\hat{\mathbf{Q}}$ to aggregate the updated embeddings, respectively. The fused embedding for each branch is then obtained by:
\begin{equation}
\begin{aligned}
\mathbf{y}=\text{AvgPooling}(\hat{\textbf{U}}),
\end{aligned}
\end{equation}
\begin{equation}
\begin{aligned}
\mathbf{z}=\text{AvgPooling}(\hat{\textbf{Q}}),
\end{aligned}
\end{equation}
where $\mathbf{y}, \mathbf{z} \in \mathbb{R}^{d}$ denote the image fused embedding and the road fused embedding, respectively.

\subsubsection{Cross-modal Constrastive Loss}

The image fused embedding $\mathbf{y}$ and the road fused embedding $\mathbf{z}$ encapsulate features across multiple modalities of the same path, reflecting an inherent similarity. Therefore, we implement a quadruplet loss function to ensure that the difference between $\mathbf{y}$ and $\mathbf{z}$ is smaller than the differences with the fused embeddings of other paths.
For negative samples, we randomly sample the image fused embedding $\mathbf{y}_\text{N}$ and the road fused embedding $\mathbf{z}_\text{N}$ from the batch. The loss function is defined as:
\begin{equation}
\begin{aligned}
\mathcal{L}_\text{fuse}&=-\sum_{p\in \mathcal{P}} ([||\mathbf{y}-\mathbf{z}||^2_2-||\mathbf{y}-\mathbf{z}_\text{N}||^2_2+\beta]_+ \\
&+[||\mathbf{y}-\mathbf{z}||^2_2-||\mathbf{z}-\mathbf{y}_\text{N}||^2_2+\beta]_+),
\end{aligned}
\end{equation}
where $\beta$ is a hyperparameter that controls the margin of the distance between pairs of positive and negative samples, and $[\cdot]_+$ is a shorthand for $\text{max}(0, \cdot )$.

\subsection{Training objective}
The final training objective of our model integrates all previously proposed loss functions,  formulated as follows:
\begin{equation}
\begin{aligned}
\mathcal{L}&=\lambda_\text{mask} \mathcal{L}_\text{mask}+\lambda_\text{multi}\mathcal{L}_\text{multi}+\lambda_\text{fuse} \mathcal{L}_\text{fuse},
\end{aligned}
\end{equation}
where $\lambda_\text{mask}$, $\lambda_\text{multi}$ and $\lambda_\text{fuse}$ are the weights assigned to $\mathcal{L}_\text{mask}$, $\mathcal{L}_\text{multi}$, and $\mathcal{L}_\text{fuse}$. 

After pre-training, we combine the image fused embedding $\mathbf{y}$ with the road fused embedding $\mathbf{z}$ into a generic path embedding $\mathbf{x}=\mathbf{y}||\mathbf{z}$, achieving a more robust and generalized representation. The generic path embedding is then fine-tuned using interchangeable linear layer task heads, enabling the model to adapt to a variety of downstream tasks effectively.

\section{Experiments}
\subsection{Experimental Setups}
\subsubsection{Datasets}
We utilize the road networks, GPS datasets, and remote sensing image datasets of two cities: Aalborg, Denmark, and Xi'an, China. The road networks are sourced from OpenStreetMap\footnote{https://www.openstreetmap.org}, while the remote sensing image datasets are acquired from Google Earth Engine\cite{gorelick2017google}. Employing an existing tool~\cite{DBLP:conf/gis/NewsonK09}, we map-match all GPS records to road networks to generate the path datasets and historical trajectory datasets. The details of the datasets are shown in Table~\ref{table:data}.

\begin{table}[!htp]
\caption{Data statistics}
\label{table:data}
\renewcommand{\arraystretch}{1.1}
\setlength{\tabcolsep}{1mm}{\begin{tabular}{c c c}
\hline
&Aalborg& Xi'an\\
\hline
Number of nodes&7,561 & 7,051 \\
Number of edges& 9,605& 9,642 \\
AVG edge length (m)&124.78 & 86.49\\
\hline
Number of path& 47,865 &200,000\\
AVG node number per road path&25.77& 55.39 \\
AVG path length (m)& 3,252.75& 4,743.70 \\
\hline
 Number of traj.& 149,246 & 797,882 \\
 Max travel time of traj. (s)& 3,549 & 8,638 \\
 Avg travel time of traj. (s)& 199 & 662 \\
\hline
Number of images& 950 & 133 \\
AVG number of nodes per image& 7.96 & 53.01\\
AVG image number per image path& 6.28 & 6.87 \\
\hline
\end{tabular}}
\end{table}

\subsubsection{Implementation Details}
All experiments are conducted using PyTorch~\cite{paszke2019pytorch} on Python 3.8 and executed on an NVIDIA Tesla-A800 GPU. Each fixed-size image is 500 $\times$ 500 pixels, with each pixel corresponding to 2 meters on the earth. In other words, an image covers a 1km $\times$ 1km region. We segment each image into 16 patches and set the embedding dimension $d$ to 64. Both the Image-Transformer and Road-Transformer comprise five layers. To enhance the pre-training efficiency, we initialize our Road-Transformer with the pre-trained LightPath~\cite{yang2023}. The mask ratio is set at 15\%. The weights $\lambda_\text{mask}$, $\lambda_\text{fuse}$, and $\lambda_\text{multi}$ are uniformly set to 1.  Training proceeds for up to 60 epochs with a learning rate of 0.02. The linear layer task head includes two fully connected layers, with dimensions of 32 and 1, respectively. Training MM-Path on the Aalborg and Xi'an datasets takes 78 and 161 minutes, respectively. Since training is conducted offline, the runtime is acceptable.

\subsubsection{Downstream Tasks and Metrics}
\textbf{Path Travel Time Estimation:} We calculate the average travel time (in seconds) for each path based on historical trajectories.
The accuracy of travel time estimations is evaluated using three metrics: Mean Absolute Error (MAE), Mean Absolute Relative Error (MARE), and Mean Absolute Percentage Error (MAPE). 
\textbf{Path Ranking Score Estimation (Path Ranking):} Each path is assigned a ranking score ranging from 0 to 1, derived from historical trajectories by following existing studies~\cite{yang2020context, yang2020learning, yang2023}. We evaluate the effectiveness of path ranking using MAE, the Kendall rank correlation coefficient ($\tau$), and Spearman’s rank correlation coefficient ($\rho$). 

\begin{table*}[!htp]
\centering
\small
\caption{Overall accuracy on travel time estimation and path ranking}
\label{tab:baseline}
\renewcommand{\arraystretch}{1.1}
\begin{tabular}{ | l |c| c| c| c| c |c |c| c| c| c| c |c | }
\hline
 \multirow{3}*{Methods} & \multicolumn{6}{c|}{Aalborg}& \multicolumn{6}{c|}{Xi'an} \\
\cline{2-13}
 \multirow{3}*{} &\multicolumn{3}{c|}{Travel Time Estimation} & \multicolumn{3}{c|}{Path Ranking}   &\multicolumn{3}{c|}{Travel Time Estimation} & \multicolumn{3}{c|}{Path Ranking}  \\
\cline{2-13}
  \multirow{3}*{} &MAE $\downarrow$&MARE $\downarrow$&MAPE $\downarrow$&MAE $\downarrow$&$\tau$  $\uparrow$&$\rho$ $\uparrow$&MAE $\downarrow$&MARE $\downarrow$&MAPE $\downarrow$&MAE $\downarrow$&$\tau$ $\uparrow$&$\rho$ $\uparrow$\\
\hline
Node2vec~\cite{grover2016node2vec} &76.228&0.281&54.182&0.203&0.119 &0.140
 &227.129&0.269&30.919&0.218&0.079 &0.098\\
PIM~\cite{yang2022} &63.812 & 0.237 & 47.054 & 0.144 & 0.284 & 0.343 & 207.266 & 0.246 & 27.716 & 0.207 & 0.091 & 0.102\\
Lightpath~\cite{yang2023} &58.818 &0.221 &40.219 & 0.124 & 0.413 & 0.483 & 201.400 & 0.229 & 26.429 & 0.178 & 0.209 & 0.252
\\
TrajCL~\cite{chang2023contrastive} &53.822 & 0.208 & 34.239 & \underline{0.113} & 0.499 & 0.577 & 202.757 & 0.238 & 26.506 & 0.181 & 0.211 & 0.256\\
{START}~\cite{jiang2023start} &\underline{51.176}&0.191&34.315&0.117&0.475 &0.556 & \underline{199.843} &0.215& \underline{25.022}&0.179&\underline{0.229} &\underline{0.279}\\
\hline
{CLIP}~\cite{radford2021learning} &72.155&0.261&50.284&0.162&0.179 &0.185 &219.048& 0.256&30.962&0.213&0.087&0.099\\
{USPM}~\cite{chen2024profiling}&66.714& 0.249 & 51.916 & 0.148 &  0.308 & 0.383  &205.594 & 0.244 & 26.039 &  0.209 & 0.105 &  0.110  \\
{JGRM}~\cite{ma2024more} &51.251&0.193&\underline{32.380}&0.115&\underline{0.512}&\underline{0.592}&201.010&0.228&26.400&\underline{0.177}&0.228&0.262\\
{Lightpath+image} &59.698 & 0.224 & 40.920 & 0.131 & 0.383 & 0.405 & 205.556 & 0.242 & 27.058 & 0.182 & 0.188 & 0.231\\
{START+image} &51.859 & \underline{0.188} & 33.401 & 0.122 & 0.437 & 0.521 & 200.059 & \underline{0.211} & 26.046 & 0.184 & 0.183 & 0.226\\
\hline
\textbf{MM-Path}&\textbf{47.756}&\textbf{0.172}&\textbf{29.808}&\textbf{0.106}&\textbf{0.558} &\textbf{0.643}&\textbf{187.452}&\textbf{0.193}&\textbf{23.644}&\textbf{0.165}&\textbf{0.257} &\textbf{0.294}\\
\hline
Improvement&6.682\% & 9.947\% & 12.941\% & 6.194\% & 11.823\% & 11.443\% & 6.201\% & 10.236\% & 5.507\% & 7.303\% & 12.227\% & 5.376\%\\
Improvement*&6.819\% & 8.511\% & 7.943\% & 7.826\% & 8.984\% & 8.614\% & 6.312\% & 8.531\% & 9.222\% & 6.780\% & 12.719\% & 12.213\%\\
\hline
\end{tabular}
\end{table*}

\subsubsection{Baselines} 
We compare the proposed model with 5 unsupervised single-modal path pre-trained methods and 5 unsupervised multi-modal methods. The single-modal path pre-trained methods are:
\begin{itemize}
    \item \textbf{Node2vec}~\cite{grover2016node2vec}: This is an unsupervised model that learn node representation based on a graph.
    \item \textbf{PIM}~\cite{yang2022}: This is an unsupervised path representation learning approach based on mutual information maximization.
    \item \textbf{LightPath}~\cite{yang2023}: This is a lightweight and scalable path representation learning method.
    \item \textbf{TrajCL}~\cite{chang2023contrastive}: This is a contrastive learning-based trajectory modeling method.
    \item \textbf{START}~\cite{jiang2023start}: This is a self-supervised trajectory representation learning framework with temporal regularities and travel semantics.
\end{itemize} 

The multi-modal methods are:
\begin{itemize}
    \item \textbf{CLIP}~\cite{radford2021learning}: This is a classic pre-trained multi-modal model. For each path, we use a single rectangular image for the image modality and replace the original text sequence with a node sequence. After pre-training, we concatenate the representations of the two modalities and use them as input to the linear layer task head.
    \item \textbf{USPM}~\cite{chen2024profiling} utilizes both images and road network to profile individual streets, but not paths (i.e., sequences of streets). We adapt USPM to support path representation learning.
    \item \textbf{JGRM}~\cite{ma2024more}: This is a representation learning framework that combines GPS data and road network-based data. In this study, we replace the GPS data with the image path.
    \item \textbf{LightPath+image}: This is a multi-modal variant of LightPath. We concatenate the patch embedding with the node embedding to replace the original node embedding for training.
    \item \textbf{START+image}: This is a multi-modal variant of START, processed similarly to LightPath+image.
\end{itemize}

For all methods, we standardize the embedding dimensionality ($d$) to 50. All parameters are set according to the specifications in the original papers.
All baselines are fine-tuned using a linear layer task head. The output of this task head serves as the prediction result. 

For all methods, we initially pre-train using unlabeled training data (e.g., 30K unlabeled Aalborg dataset and 160K unlabeled Xi'an dataset). Subsequently, we use a smaller volume of labeled data (e.g., 10K labeled Aalborg dataset and 40K labeled Xi'an dataset) for task-specific fine-tuning. Validation and evaluation processes are conducted on separate validation dataset (e.g., 5K Aalborg dataset and 20K Xi'an dataset) and test dataset (e.g., 10K Aalborg dataset and 40K Xi'an dataset), respectively.

\subsection{Experimental Results}
\subsubsection{Overall Performance}
Table~\ref{tab:baseline} presents the overall performance on both tasks. We use ‘$\uparrow$’ (and ‘$\downarrow$’) to indicate that larger (and smaller) values are better. For each task, we highlight the best and second-best performance in \textbf{bold} and \underline{underline}. ‘‘Improvement’’ and ‘‘Improvement*’’  quantify the enhancements achieved by MM-Path over the best single-modal and multi-modal baselines, respectively.

Overall, MM-Path outperforms all baselines on these tasks across both datasets, demonstrating its superiority. Specifically, we can make the following observations: The graph representation learning method Node2vec significantly underperforms compared to MM-Path, primarily  due to its focus solely on the topological information of nodes while overlooking the sequential information of paths.  Single-modal models like PIM, LightPath, and TracjCL show improved performance over Node2vec, indicating the importance of capturing sequential correlations within paths. Among the single-modal models, START achieves the best performance. It adeptly integrate sequential path information with spatio-temporal transition relationships derived from historical trajectory data. However, as a single-modal model, its capabilities are inherently constrained. As a multi-modal model, CLIP exhibits the weakest performance. Designed primarily for general corpora, it focuses on single, coarse-grained image representations, which often introduce noise into path modeling. Consequently, CLIP struggles to effectively capture complex spatial information and correspondences, making it unsuitable for modeling paths. USPM performs poorly because it analyzes individual streets using images and road networks, rather than paths (i.e., street sequences). As a result, it fails to effectively mine the sequential relationships present in the two modalities.
The variants LightPath+image and START+image perform comparably to their single-modal models (i.e., LightPath and START), suggesting that merely concatenating two modalities does not effectively enhance multi-modal fusion. 
Having adapted JGRM to integrate image paths and road paths, JGRM outperforms other multi-modal baselines. It is specifically designed for multi-modal integration and excels at merging information from various sources. However, JGRM's limitations in handling multi-modal information of varying granularities and its lack of use of cross-modal context information to guide the fusion process make its performance less optimal compared to MM-Path.

\subsubsection{Ablation Study}
We design eight variants of MM-Path to verify the necessity of the components of our model:
(1) MM-Path-y: This model utilizes the fused embedding $\mathbf{y}$  (cf. Eq. 10)  as a generic representation of the path.
(2) MM-Path-z: This variant leverages the fused embedding $\mathbf{z}$ (cf. Eq. 11) as a generic representation of the path.
(3) w/o alignment: This version excludes the multi-granularity loss.
(4) w/o fusion: This variant substitutes  the graph-based residual fusion component with average pooling of the encoded embeddings from both modalities.
(5) w/o GCN: This model replaces the GCN in the graph-based cross-modal residual fusion component with a cross-attention mechanism. 
(6) w/o fine, (7) w/o medium, and (8) w/o coarse: These variants omit the fine-grained, medium-grained, and coarse-grained loss, respectively.

The results are summarized in Tables~\ref{tab:ablation1} and~\ref{tab:ablation2}. We can observe that MM-Path w/o alignment shows poor performance, which is attributed to its reliance solely on multi-modal data fusion without considering multi-granularity alignment. The variants, w/o fine, w/o medium, and w/o coarse, outperform w/o alignment but still worse than the full MM-Path, demonstrating the importance of multiple granularity alignments. MM-Path w/o fusion also exhibits poor performance, while MM-Path w/o GCN performs slightly worse than MM-Path. These results indicate that complex fusion methods with cross-modal context information enhance path understanding. Both MM-Path-y and MM-Path-z demonstrate comparable performance in travel time estimation and path ranking. This indicates that different modal perspectives contribute valuable insights for various downstream tasks. The overall performance of MM-Path surpasses all variants. This result implies that each of the proposed components significantly enhances the model's effectiveness. This conclusively validates that MM-Path optimally utilizes all designed components.

\begin{table}[!ht]
\centering
\small
\caption{Effect of variants of MM-Path in Aalborg}
\vspace{-1em}
\label{tab:ablation1}
\renewcommand{\arraystretch}{1.02}
\setlength{\tabcolsep}{1.5mm}{
\begin{tabular}{ | l |c| c| c| c| c |c |}
\hline
 \multirow{3}*{Methods} & \multicolumn{6}{c|}{Aalborg}\\
\cline{2-7}
 \multirow{3}*{} &\multicolumn{3}{c|}{Travel Time Estimation} & \multicolumn{3}{c|}{Path Ranking}\\
\cline{2-7}
 \multirow{3}*{} &MAE&MARE&MAPE&MAE&$\tau$ &$\rho$\\
\hline
MM-Path-y &48.529&0.181&32.722&0.118&0.511&0.603
\\
MM-Path-z& 49.649&0.185&30.193&0.114&0.528&0.622
\\
w/o alignment &52.832&0.201&36.251&0.131&0.300&0.379
\\
w/o fusion &51.237&0.192&30.529&0.115&0.476&0.560
\\
w/o GCN &48.651&0.183&33.371&0.111&0.532&0.619
\\
w/o fine&51.641&0.192&33.277&0.129&0.441&0.523
\\
w/o medium&50.932&0.187&34.250&0.114&0.494& 0.583
\\
w/o coarse&50.688&0.189&35.341&0.117&0.505&0.596\\
\hline
\textbf{MM-Path}&\textbf{47.756}&\textbf{0.172}&\textbf{29.808}&\textbf{0.106}&\textbf{0.558} &\textbf{0.643}\\
\hline
\end{tabular}}
\end{table}

\begin{table}[!ht]
\centering
\small
\caption{Effect of variants of MM-Path in Xi'an}
\vspace{-1em}
\label{tab:ablation2}
\renewcommand{\arraystretch}{1.02}
\setlength{\tabcolsep}{1.5mm}{
\begin{tabular}{ | l  |c| c| c| c| c |c | }
\hline
 \multirow{3}*{Methods} & \multicolumn{6}{c|}{Xi'an}\\
\cline{2-7}
 \multirow{3}*{} &\multicolumn{3}{c|}{Travel Time Estimation} & \multicolumn{3}{c|}{Path Ranking}\\
\cline{2-7}
 \multirow{3}*{} &MAE&MARE&MAPE&MAE&$\tau$ &$\rho$\\
\hline
MM-Path-y &196.331&0.233& 24.747&0.196&0.178&0.214
\\
MM-Path-z& 194.301&0.231&24.455& 0.183&0.199&0.241
\\
 w/o alignment &200.335&0.239&26.459&0.195&0.131&0.167

\\
w/o fusion &200.652&0.239&25.433&0.208&0.113&0.130
\\
w/o GCN &189.659&0.221&24.496&0.173&0.234&0.286
\\
w/o fine&199.214& 0.235& 26.913&0.177&0.226&0.275
\\
w/o medium&192.514&0.229&24.826&0.175&0.227&0.278
\\
w/o coarse&194.256&0.230&25.757&0.176&0.231&0.278\\
\hline
\textbf{MM-Path}&\textbf{187.452}&\textbf{0.193}&\textbf{23.644}&\textbf{0.165}&\textbf{0.257} &\textbf{0.294}\\
\hline
\end{tabular}}
\end{table}

\begin{figure*}[!ht]
    \centering
    \subfigure[Travel Time Estimation, Aalborg]{
    \includegraphics[width=0.22\textwidth]{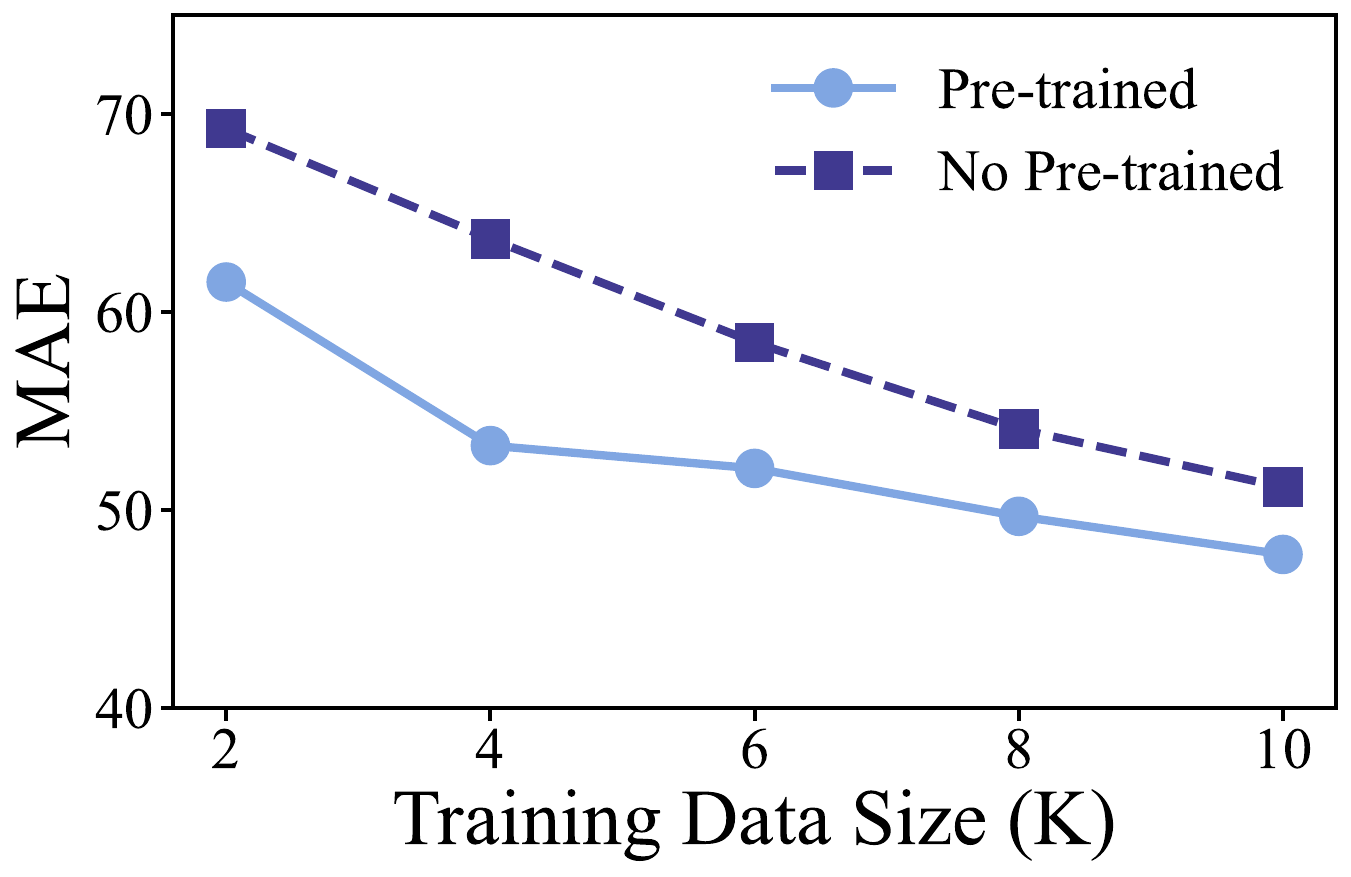}
    }
    \subfigure[Path Ranking, Aalborg]{
    \includegraphics[width=0.22\textwidth]{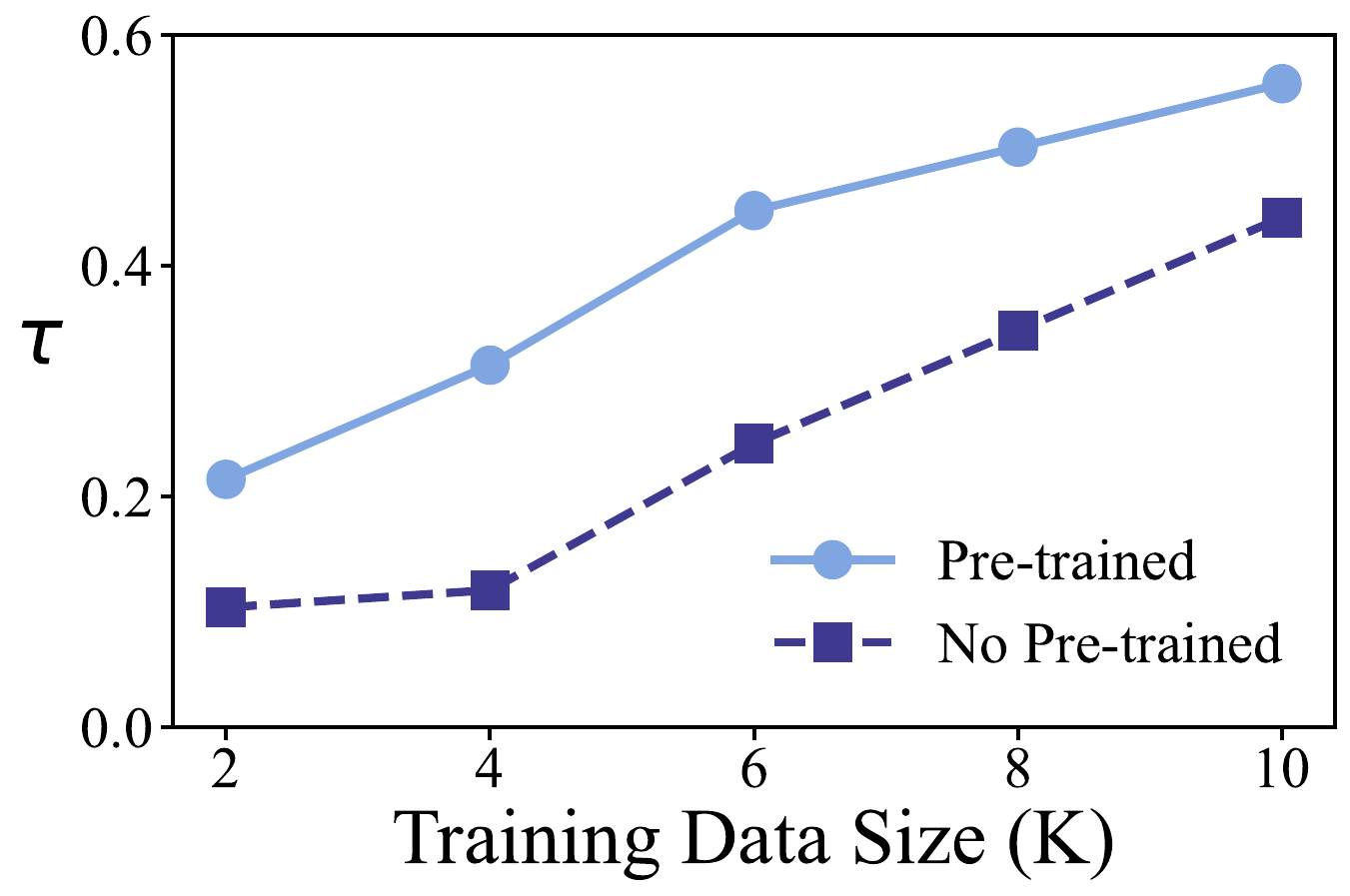}
    }
    \subfigure[Travel Time Estimation, Xi'an]{
        \includegraphics[width=0.22\textwidth]{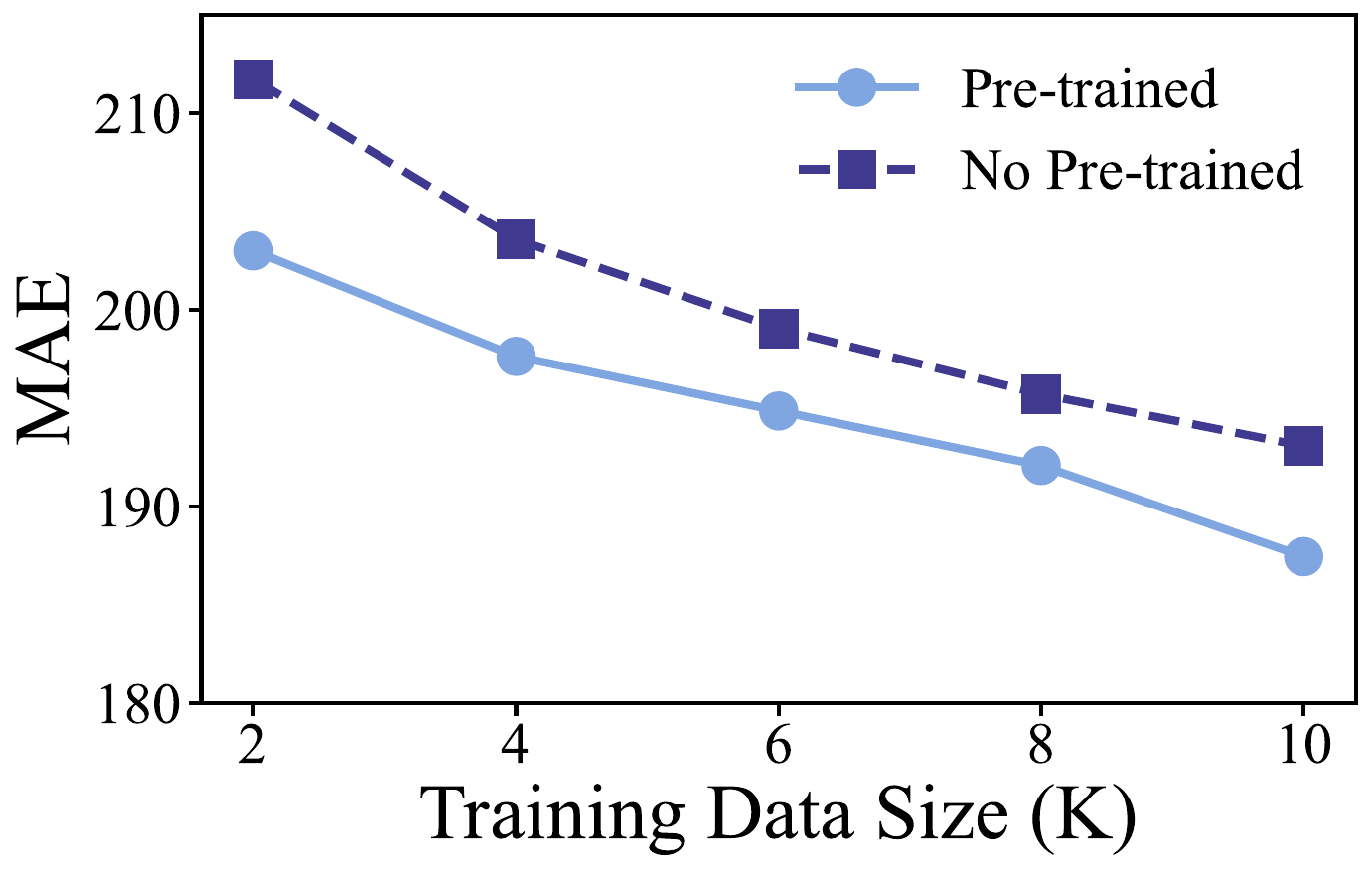}
    }
    \subfigure[Path Ranking, Xi'an]{
        \includegraphics[width=0.22\textwidth]{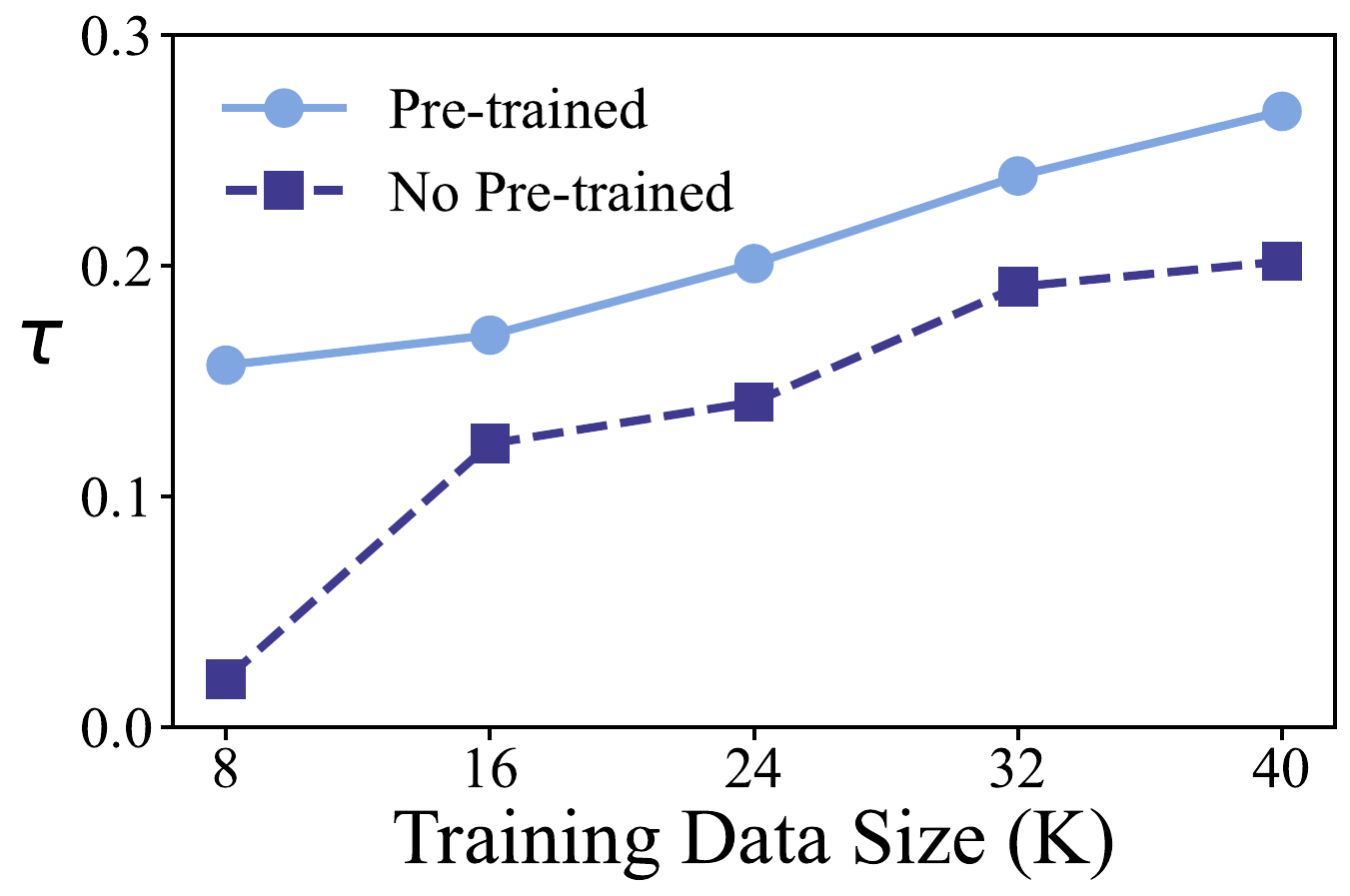}
    }
    \vspace{-1em}
    \caption{Effect of pre-training}
    \vspace{-1em}
    \label{fig:pre-train}
\end{figure*}

\begin{figure*}[!ht]
    \centering
    \subfigure[Travel Time Estimation, Aalborg]{
    \includegraphics[width=0.22\textwidth]{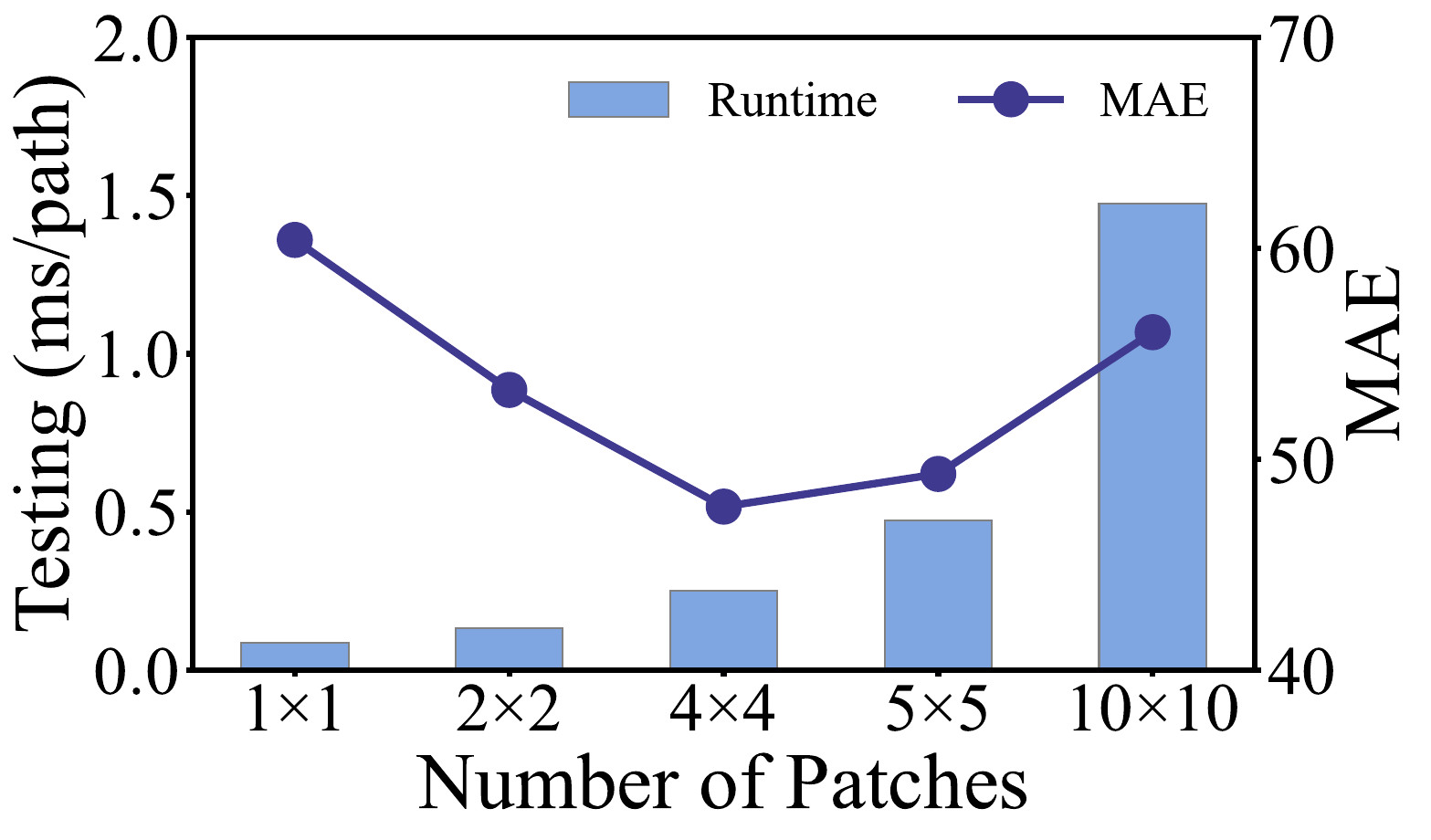}
    }
    \subfigure[Path Ranking, Aalborg]{
    \includegraphics[width=0.22\textwidth]{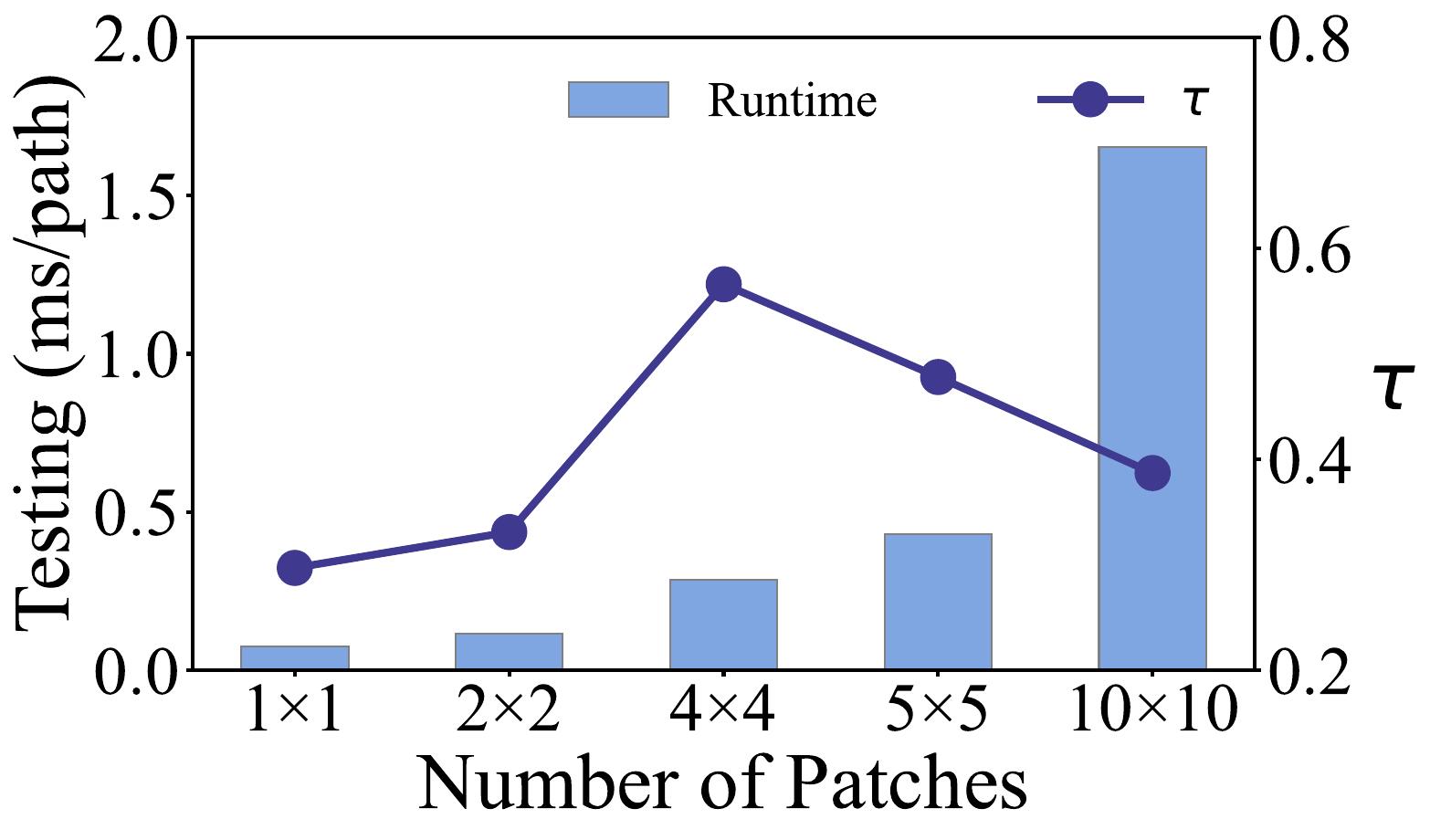}
    }
    \subfigure[Travel Time Estimation, Xi'an]{
    \includegraphics[width=0.22\textwidth]{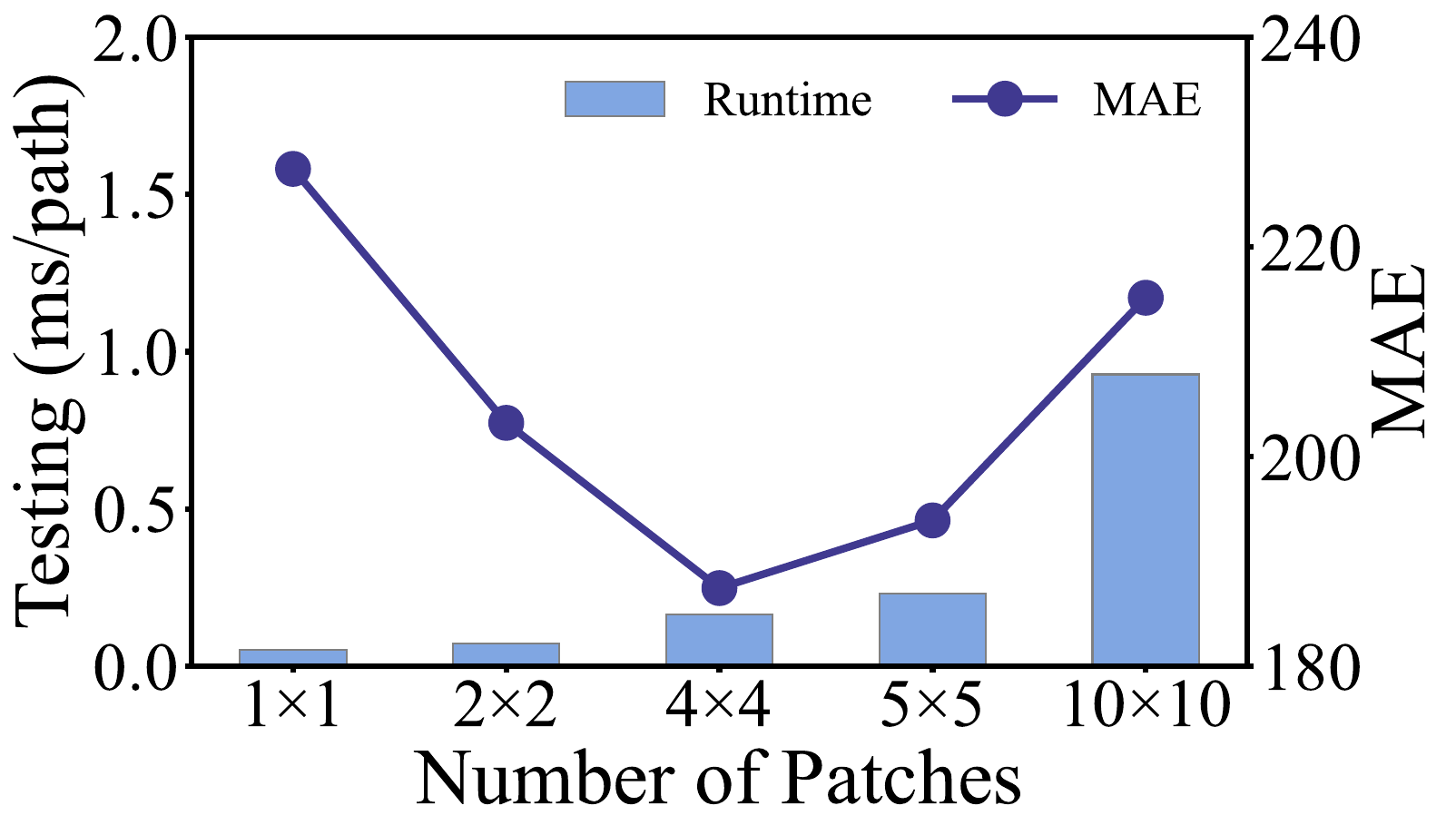}
    }
    \subfigure[Path Ranking, Xi'an]{
    \includegraphics[width=0.22\textwidth]{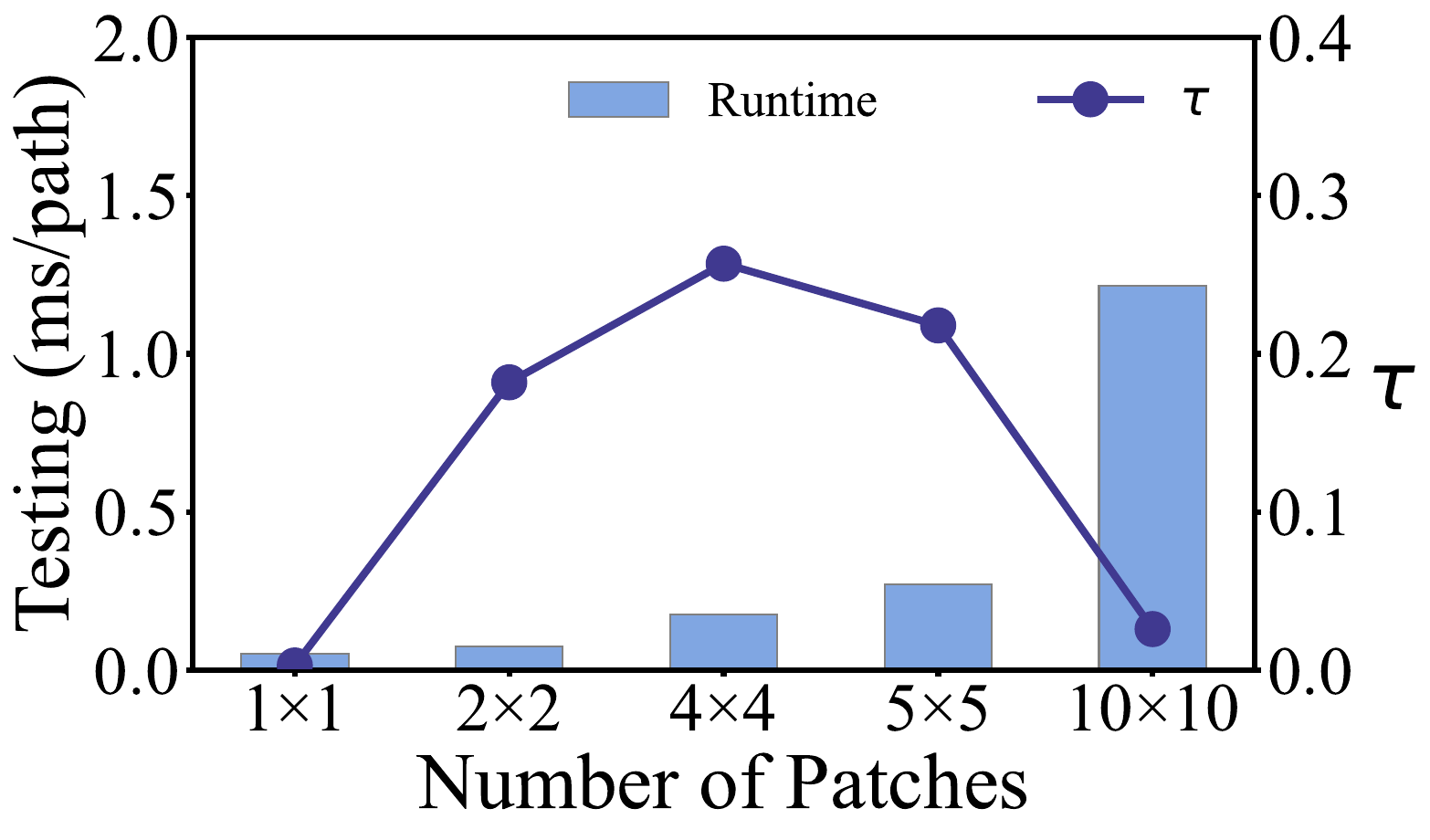}
    }
    \vspace{-1em}
    \caption{Effect of varying granularity size}
    \label{fig:patch_aal}
\end{figure*}

\begin{table*}[!htp]
\centering
\small
\caption{Accuracy on travel time estimation and path ranking for short paths}
\vspace{-1em}
\label{tab:short}
\renewcommand{\arraystretch}{1.07}
\begin{tabular}{ | l |c| c| c| c| c |c |c| c| c| c| c |c | }
\hline
 \multirow{3}*{Methods} & \multicolumn{6}{c|}{Aalborg}& \multicolumn{6}{c|}{Xi'an} \\
\cline{2-13}
 \multirow{3}*{} &\multicolumn{3}{c|}{Travel Time Estimation} & \multicolumn{3}{c|}{Path Ranking}   &\multicolumn{3}{c|}{Travel Time Estimation} & \multicolumn{3}{c|}{Path Ranking}  \\
\cline{2-13}
 \multirow{3}*{} &MAE $\downarrow$&MARE $\downarrow$&MAPE $\downarrow$&MAE $\downarrow$&$\tau$  $\uparrow$&$\rho$ $\uparrow$&MAE $\downarrow$&MARE $\downarrow$&MAPE $\downarrow$&MAE $\downarrow$&$\tau$ $\uparrow$&$\rho$ $\uparrow$\\
\hline
Node2vec~\cite{grover2016node2vec} &67.898&0.306&55.179&0.201&0.140&0.153
 &179.828&0.269&32.512&0.220&0.081 &0.084\\
PIM~\cite{yang2022} &55.871 & 0.252 & 44.605 &  0.150 &  0.292 & 0.348 &  150.356 & 0.238 & 26.221 & 0.206 & 0.174 & 0.198\\
Lightpath~\cite{yang2023} &52.708 &0.237 & 40.414 &  0.127 & 0.445 & 0.502 &149.816 & 0.237 &26.702 & 0.180 & 0.201 & 0.249
\\
TrajCL~\cite{chang2023contrastive} &49.427 & 0.223 & 34.466 & \underline{0.116} & \underline{0.523} & 0.597 & 150.657 & 0.239 &  26.891 & 0.183 & 0.219 & 0.260\\
{START}~\cite{jiang2023start} & \underline{47.347}&0.216&36.523&0.122&0.513&0.586 & \underline{147.745} &\underline{0.227}& \underline{25.973}&0.179&0.219 &\underline{0.277}\\
\hline
{CLIP}~\cite{radford2021learning} &62.300& 0.281& 49.164&0.166&0.176&0.217&170.818&0.271&33.201&0.215&0.079&0.095\\
{USPM}~\cite{chen2024profiling} &57.415&0.274& 48.949&0.152&0.314&0.360&149.012&0.234&26.942&0.207&0.133&0.151\\
{JGRM}~\cite{ma2024more} &47.770&\underline{0.213}&\underline{33.715}&0.117&0.516&\underline{0.612}&148.832&0.236&26.341&\underline{0.176}&\underline{0.221}&0.262\\
{Lightpath+image} &54.819 &  0.247 &  42.258 & 0.134 & 0.383 & 0.434 & 150.657 &0.239 & 26.891 & 0.184 & 0.186 & 0.212\\
{START+image} &49.427 &  0.223 & 34.462 &  0.125 & 0.476 & 0.548& 148.684 &  0.236 &26.503 & 0.185 & 0.181 & 0.231\\
\hline
\textbf{MM-Path}&\textbf{44.092}&\textbf{0.202}&\textbf{31.058}&\textbf{0.110}&\textbf{0.562} &\textbf{0.662}&\textbf{138.543}&\textbf{ 0.218}&\textbf{25.112}&\textbf{0.164}&\textbf{0.253} &\textbf{0.294}\\
\hline
Improvement&6.874\% & 6.481\% & 9.809\% & 5.172\% & 7.456\% & 10.887\% & 6.228\% & 

3.964\% & 

3.315\% &8.379\% & 15.525\% & 5.755\%\\
Improvement*&7.699\% & 5.164\% & 7.881\% & 5.982\% & 8.914\% & 8.169\% & 

6.821\% & 7.234\% & 4.665\% & 6.818\% & 14.479\% & 12.213\%\\
\hline
\end{tabular}
\end{table*}

\begin{table*}[!ht]
\centering
\small
\caption{Accuracy on travel time estimation and path ranking for long paths}
\vspace{-1em}
\label{tab:long}
\renewcommand{\arraystretch}{1.07}
\begin{tabular}{ | l |c| c| c| c| c |c |c| c| c| c| c |c | }
\hline
 \multirow{3}*{Methods} & \multicolumn{6}{c|}{Aalborg}& \multicolumn{6}{c|}{Xi'an} \\
\cline{2-13}
 \multirow{3}*{} &\multicolumn{3}{c|}{Travel Time Estimation} & \multicolumn{3}{c|}{Path Ranking}   &\multicolumn{3}{c|}{Travel Time Estimation} & \multicolumn{3}{c|}{Path Ranking}  \\
\cline{2-13}
 \multirow{3}*{} &MAE $\downarrow$&MARE $\downarrow$&MAPE $\downarrow$&MAE $\downarrow$&$\tau$  $\uparrow$&$\rho$ $\uparrow$&MAE $\downarrow$&MARE $\downarrow$&MAPE $\downarrow$&MAE $\downarrow$&$\tau$ $\uparrow$&$\rho$ $\uparrow$\\
\hline
Node2vec~\cite{grover2016node2vec} &193.021&0.295&25.416&0.231& 0.019 &-0.008
 &262.160&0.279&28.294&0.171&0.015&0.108\\
PIM~\cite{yang2022} &135.348 &  0.207 & 17.855 & 0.098 & 0.210 & 0.239 & 246.606 & 0.246 & 24.981 &0.186 & 0.179 & 0.207\\
Lightpath~\cite{yang2023} &114.599 & 0.175 & 16.079 & 0.080 & 0.371 & 0.326 & 241.666 &  0.237 & 24.752 & \underline{0.107} & 0.259 & 0.317
\\
TrajCL~\cite{chang2023contrastive} &99.041 & 0.151 & 14.215 & 0.077 &0.346 & 0.372 &  243.581 & 0.238 &24.951 & 0.111 &\underline{0.266} &\underline{0.328}\\
{START}~\cite{jiang2023start} & \underline{87.890}&\underline{0.136}&13.220&0.078&0.328 &0.364 & \underline{237.245} &\underline{0.230}& \underline{23.973}&0.108&0.244 &0.301\\
\hline
{CLIP}~\cite{radford2021learning} & 174.421&0.266& 23.065&0.126&0.145 &0.172 &255.160&0.249&28.752&0.208&0.093&0.104\\

{USPM}~\cite{chen2024profiling} &147.236&	0.232&	24.435&		0.105&	0.201&	0.218&243.380&	0.240&	24.549&	0.157&	0.216&	0.277\\

{JGRM}~\cite{ma2024more} &89.125&0.138&\underline{13.131}&\underline{0.076}&\underline{0.381}&\underline{0.379}&239.386&0.234&24.379&0.109&0.265&0.319\\
{Lightpath+image} &111.703 &0.171 & 15.897 & 0.092 & 0.326 & 0.308 & 243.581 & 0.238 & 24.951 & 0.112 & 0.236 & 0.323\\
{START+image} & 99.041 & 0.151 & 14.215 & 0.083 & 0.317 & 0.323 & 237.347 & 0.232 & 24.641 & 0.113 & 0.258 & 0.312\\
\hline
\textbf{MM-Path}&\textbf{83.125}&\textbf{0.121}&\textbf{11.214}&\textbf{0.067}&\textbf{0.421} &\textbf{0.424}&\textbf{224.227}&\textbf{0.190}&\textbf{21.832}&\textbf{0.103}&\textbf{0.279} &\textbf{0.345}\\
\hline
Improvement&5.421\% & 11.029\% & 15.174\% & 12.987\% & 13.477\% &13.978\% &3.833\% & 17.391\% & 8.931\% & 3.738\% & 2.307\% & 5.182\%\\
Improvement*&6.732\% &12.318\% & 14.599\% &11.842\% & 10.499\% &

11.873\% & 5.527\% & 18.103\% & 10.447\% & 5.504\% & 5.283\% & 7.836\%\\
\hline
\end{tabular}
\end{table*}

\subsubsection{Effect of Pre-training}
In this section, we evaluate the effect of pre-training. We vary the size of labeled data used for fine-tuning, and compare the performance of the proposed MM-Path (Pre-trained) with its variant that lacks  pre-training (No Pre-trained). Figure~\ref{fig:pre-train} shows the performance of travel time estimation and path ranking. We observe that the performance of both models improves with an increase in labeled data size, and the pre-trained model consistently outperforms the no pre-trained model. This illustrates that the pre-trained model, equipped with extensive cross-modal context information, requires less labeled data and achieves superior performance compared to the model without pre-training. These findings suggest that MM-Path can effectively serve as a pre-training model to enhance supervised learning methods.

\subsubsection{Parameter Sensitivity}
We explore the impact of image granularity size on the model's performance. For uniform segmentation,  each 500 $\times$ 500 pixel image is segmented into 1$\times$1, 2$\times$2, 4$\times$4, 5$\times$5, and 10$\times$10 patches, respectively. The performance and testing runtime for each granularity size are detailed in Figure~\ref{fig:patch_aal}. 

We observe that a larger number of patches also incurs longer inference runtime for each path. Moreover, the model's performance improves as the number of patches increases from 1$\times$1 to 4$\times$4, suggesting that finer granularity enhances the capture of detailed features. However, performance begins to decline with further increases to 5$\times$5 and 10$\times$10 patches. This decrease is due to the limited context features extracted by excessively fine granularity, which negatively impacts path understanding.

Then, we explore the model scalability in terms of road path length. We evaluate the performance of all models on paths with varying numbers of nodes.  Specifically, paths with fewer than 50 nodes are classified as short paths, while those with more than 50 nodes are classified as long paths. The performance of all models on short and long paths is detailed in Tables~\ref{tab:short} and~\ref{tab:long}, respectively. The experimental results show that MM-Path outperforms other models across both path lengths, demonstrating its superiority.

\subsubsection{Case Study}
\begin{figure}[!ht]
	\begin{center}
    \includegraphics[width=0.9\linewidth]{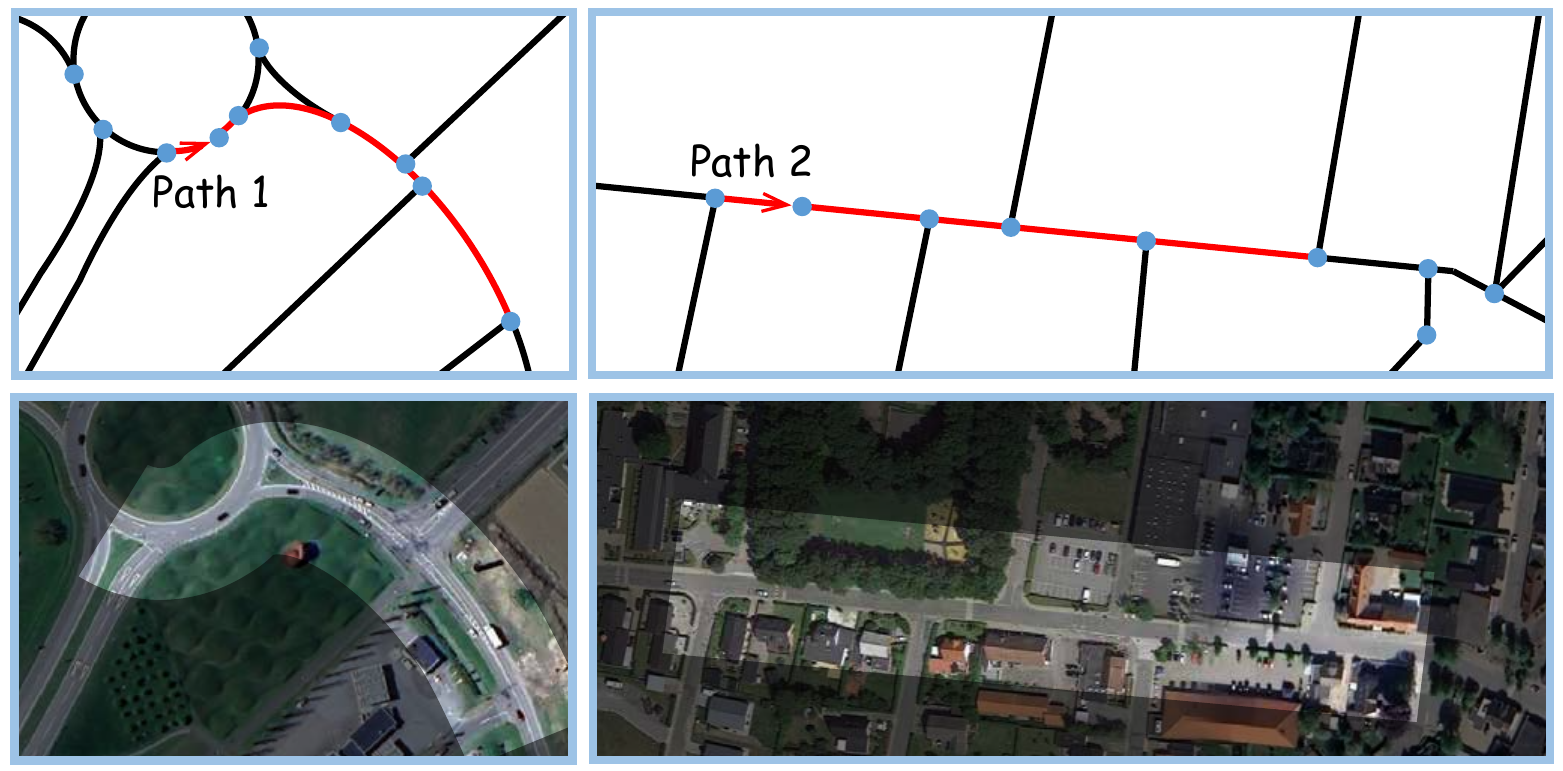}
	\end{center}
	\caption{Visualization of two paths}
	\label{fig:case}
\end{figure}

We inspect a pair of representative paths in Aalborg to demonstrate the superiority of MM-Path. The road paths and image paths are visualized in Figure~\ref{fig:case}. The travel time estimation results of MM-Path and the superior baselines are shown in Table~\ref{tab:case}.

Two paths in Figure~\ref{fig:case} exhibit a similar structure on the road network, both having a node degree sequence of $\langle 3, 2, 3, 3, 3\rangle$. Such single-modal data might suggest that these paths have comparable travel times. However, the visual information from their images differs significantly. Specifically, path 1 traverses a roundabout and runs along a trunk road, where paths typically allow for faster travel speeds. In contrast, path 2 is located on an ordinary road near residential buildings, typically associated with slower speeds. 

As shown in Table~\ref{tab:case}, the travel time estimates from TrajCL, START, and START+image suggest a shorter travel time for path 2 and a longer travel time for path 1, which are contrary to the ground truth.
This illustrates that a single modality provides limited information, and the simple concatenation of multi-modal data in the START+image model fails to effectively extract image information. In contrast, the results from the multi-modal model JGRM align with the relative magnitudes of the actual travel times, but its estimates exhibit large deviations. Meanwhile, MM-Path demonstrates superior travel time estimation performance compared to other models, indicating its effective fusion and utilization of image information.

\begin{table}[!ht]
\centering
\small
\caption{Travel time estimation results of different models}
\renewcommand{\arraystretch}{1.1}
\setlength{\tabcolsep}{1mm}{
\begin{tabular}{ c | c | c c c c c}
\bottomrule
\multirow{2}*{Case}&Ground&\multicolumn{5}{c}{Travel Time Estimation}\\
\cline{3-7}
 \multirow{2}*{}&truth& MM-Path & TrajCL & START & JGRM & START+image\\
\hline 
Path 1& 17.0& 22.7 & 30.0 & 32.9 & 26.9 & 29.2\\
Path 2&36.0&40.1&17.1&24.3&73.8&26.5\\
\toprule
\end{tabular}}
\label{tab:case}
\end{table}

\section{Related work}

\subsection{Path Representation Learning Models}

With the advancement of location-based services, the trajectory generation~\cite{ zhu2023difftraj, zhu2024controltraj} and trajectory analysis~\cite{lun2024resisting, liu2024lighttr, chen2023, ding2018ultraman, chen2019real} have become increasingly prevalent. To better understand static representation of trajectories, recent studies have increasingly focused on developing path representation learning models that do not rely on labeled training data, showing robust generalization across multiple downstream tasks~\cite{yang2023, ma2024more, jiang2023start,chang2023contrastive}. 
For instance, Jiang et al.~\cite{jiang2023start} introduce a self-supervised trajectory representation learning framework that includes tasks such as span-masked trajectory recovery and trajectory contrastive learning to leverage temporal patterns and travel semantics effectively. 
Yang et al.~\cite{yang2023} aim to minimize resource consumption and enhance model scalability by developing LightPath, a lightweight and scalable framework designed to conserve resources while maintaining accuracy. 
Additionally, Ma et al.~\cite{ma2024more} propose a representation learning framework that integrates GPS and route modeling based on self-supervised technology, further expanding the field's methodologies.

Recent advancements in Large Language Models (LLMs)~\cite{tay2022ul2,thoppilan2022lamda} have facilitated the development of general spatio-temporal prediction models~\cite{yuan2024unist, li2024urbangpt, li2024flashst, kieu2024Team, DBLP:journals/pvldb/ZhaoGCHZY23}. For example, Unist~\cite{yuan2024unist} maps spatio-temporal data onto grids and provides generic predictions across various scenarios using elaborated masking strategies and spatio-temporal knowledge-guided prompts. However, since each path exhibits geographical continuity and cannot be discretized into a single region, these models are not well-suited for path modeling. 

Despite recent advancements, current path representation learning models overlook the potential contributions of images in path understanding, which can provide valuable insights into the geometric features and contextual environmental information from a global perspective.

\subsection{Multi-modal Pre-trained Models}
Multi-modal pre-trained models aim to enhance target representation by leveraging diverse modalities, including text, images, and audio. Some methodologies converge the information from various modalities into a unified latent space
~\cite{wang2020urban2vec, radford2021learning}. For example, Wang et al.~\cite{wang2020urban2vec} propose an unsupervised multi-modal method that encodes visual, textual, and geospatial data of urban neighborhoods into a unified vector space using multiple triplet loss functions.  Radford et al.~\cite{radford2021learning} develop CLIP, a large-scale multi-modal model that evaluates the congruence of image-text pairs through cosine similarity. Some research opts to map each modality into distinct embedding spaces, while enforcing fusion on the representations~\cite{Pramanick_2023_ICCV, radford2021learning, bao2022vlmo}. For instance, Pramanick et al.~\cite{Pramanick_2023_ICCV} achieve robust video-text representation by embedding cross-modal fusion within the core video and language structures, thereby facilitating various downstream tasks and reducing fine-tuning requirements. 
Bao et al.~\cite{bao2022vlmo} present VLMO, a unified vision-language pre-trained model that jointly trains a dual encoder and a fusion encoder within a modular Transformer network.

Nevertheless, these models are predominantly trained on general corpora. The characteristics of road paths and image paths, which include complex correspondences and spatial topological relationships, are distinct from those found in conventional multi-modal datasets. Consequently, existing multi-modal models exhibit limited generalizability to this specialized domain.

\section{Conclusion and future work}
In this paper, we propose a Multi-modal Multi-granularity Path Representation Learning Framework (MM-Path), which is the first work that integrate data from road networks and remote sensing images into generic path representation learning. Initially, we model the road paths and image paths separately, and implement a multi-granularity alignment strategy to ensure the synchronization of both detailed local information and broader global context. Furthermore, we develop a graph-based cross-modal residual fusion component that effectively fuses information from both modalities while preserving the semantic consistency between modalities. MM-Path outperforms all baselines on two real-world datasets across two downstream tasks, demonstrating its superiority.

In the future, we plan to further investigate the capability of multi-modal models for generic path representing learning, with particular focus on few-shot and zero-shot learning scenarios.

\bibliographystyle{ACM-Reference-Format}
\bibliography{sample-base}

\end{document}